\documentclass[11pt]{article}

\usepackage[final]{acl}

\usepackage{times}
\usepackage{latexsym}

\usepackage[T1]{fontenc}
\usepackage[utf8]{inputenc}

\usepackage{microtype}

\usepackage{inconsolata}

\usepackage[table]{xcolor}
\usepackage{adjustbox}
\usepackage{graphicx}
\usepackage[most]{tcolorbox}
\usepackage{enumitem}

\usepackage{amsmath}
\usepackage{multirow}
\usepackage{booktabs}
\usepackage{natbib}
\usepackage{algorithm}
\usepackage{algpseudocode}
\usepackage{subcaption}
\usepackage{adjustbox}

\tcbset{
  promptbox/.style={
    enhanced,
    colback=gray!10,
    colframe=black!50,
    fontupper=\ttfamily\tiny,
    boxrule=0.4pt,
    sharp corners,
    left=6pt,
    right=6pt,
    top=4pt,
    bottom=4pt,
    boxsep=2pt,
    before skip=10pt,
    after skip=10pt,
    coltitle=black,
    attach title to upper,
    halign title=left,
    title style={font=\bfseries\tiny, colback=gray!20, colframe=black!50, boxrule=0.4pt, sharp corners},
  }
}
\usepackage{xcolor}
\usepackage{colortbl}
\definecolor{lightgreen}{HTML}{B9FAD5} % A soft, professional green
\newcommand{\cmprsr}[1]{\cellcolor{green!30}\textbf{#1}}

\usepackage{cuted}   % Allows page-wide content in two-column mode
\usepackage{caption} % Required for \captionof
\title{\centering \Large \textbf{Cmprsr}: Abstractive Token-Level Question-Agnostic Prompt Compressor}

\author{%
\parbox{\textwidth}{\centering
\vspace{10pt}
Ivan Zakazov$^{1,2}$ \quad
Berke Argın$^{1,2}$\thanks{Authors contributed equally.} \quad
Oussama Gabouj$^{1,2}$\footnotemark[\value{footnote}] \quad
Kamel Charaf$^{1,2}$\footnotemark[\value{footnote}] \\
Alexander Sharipov$^{2}$\footnotemark[\value{footnote}] \quad
Alexi Semiz$^{2}$ \quad
Lorenzo Drudi$^{2}$ \quad
Nicolas Baldwin$^{2}$ \\
Robert West$^{2}$\\
\vspace{6pt}
$^{1}$Compresr Inc. \quad $^{2}$EPFL\\
\vspace{6pt}
\small \texttt{\{ivan, berke, oussama, kamel\}@compresr.ai}\\
\small \texttt{\{alexander.sharipov, robert.west\}@epfl.ch}\\
\vspace{30pt}
}
}

\begin{document}

\maketitle

\begin{strip}
    \vspace{-2.4cm}
    \centering
    % 1. The Centered Image
    \includegraphics[width=\linewidth]{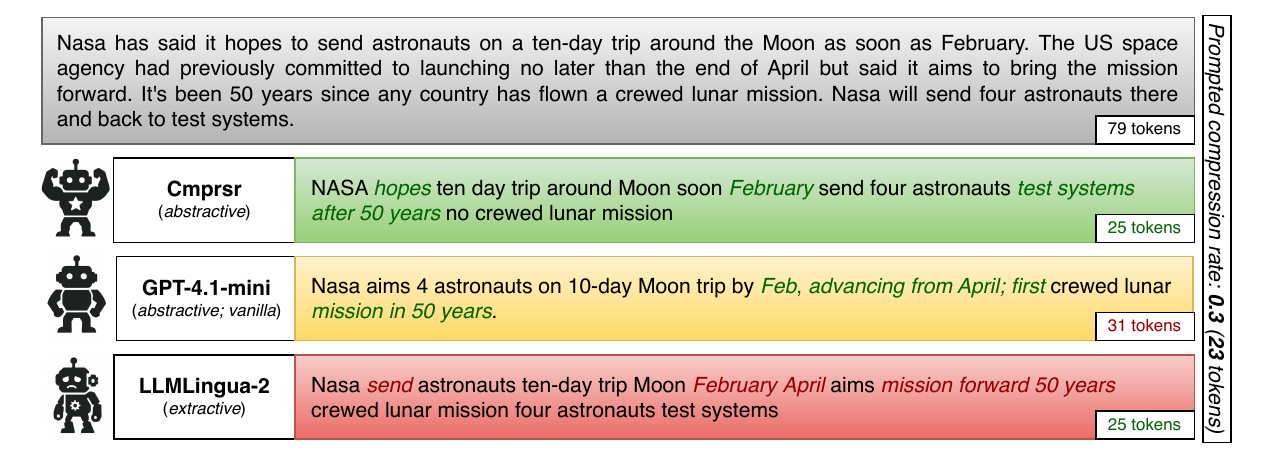}
    
    % 2. The Centered Caption
    \captionsetup{hypcap=false}
    \captionof{figure}{Extractive compression selects a subset of the input sequence tokens, while abstractive compression allows for clever paraphrases. While one can use vanilla LLMs as abstractive compressors, their performance can be further improved with RL-based post-training, yielding \textbf{Cmprsr}. Note that extractive compression may introduce ambiguities, \textit{e.g.} ``February April'', ``mission forward 50 years'' \cite{bbc2025}, while vanilla abstractive compression does not adhere to the desired compression rate.}
    \label{fig:teaser}
    
\end{strip}

% The two-column layout starts automatically here
\begin{abstract}

Motivated by the high costs of using black-box Large Language Models (LLMs), we introduce a novel prompt compression paradigm, under which we use smaller LLMs to compress inputs for the larger ones. We present the first comprehensive LLM-as-a-compressor benchmark spanning $25$ open- and closed-source models, which reveals significant disparity in models' compression ability in terms of (i) preserving semantically important information (ii) following the user-provided compression rate (CR). We further improve the performance of \textit{gpt-4.1-mini}, the best overall vanilla compressor, with \textit{Textgrad}-based compression meta-prompt optimization. We also identify the most promising open-source vanilla LLM---\textit{Qwen3-4B}---and post-train it with a  combination of \textit{supervised fine-tuning (SFT)} and \textit{Group Relative Policy Optimization (GRPO)}, pursuing the dual objective of CR adherence and maximizing the downstream task performance. We call the resulting model \textbf{Cmprsr} and demonstrate its superiority over both extractive and vanilla abstractive compression across the entire range of compression rates on lengthy inputs from \textit{MeetingBank} and \textit{LongBench} as well as short prompts from \textit{GSM8k}. The latter highlights \textbf{Cmprsr}'s generalizability across varying input lengths and domains. Moreover, \textbf{Cmprsr} closely follows the requested compression rate, offering fine control over the cost-quality trade-off. Importantly, we show that in case of lengthy inputs, compression can be practically lossless in terms of the downstream task performance.

\end{abstract}

% \section{Introduction}

% \section{Abstractive compression with LLMs}
\section{Introduction}
\label{Sec:intro}

The discovery of scaling laws \cite{kaplan2020scalinglawsneurallanguage} set the trend for training increasingly large Language Models (LMs). Despite rapid advancements in both hardware and software supporting LLMs' inference, the costs of their usage continue to surge. This trend reflects not only growing adoption \cite{liang2025widespreadadoptionlargelanguage}, but also the Jevons paradox \cite{Luccioni25}: efficiency gains that spur even greater consumption.

According to recent estimates \cite{tully2025midyear}, $87\%$ of the company's spending on LLMs is attributed to black-box LLMs, accessed via API. This means that in practice the only way the majority of LLM consumers can optimize their spendings is through minimizing the queries' length passed to the models; this can be achieved with the \textit{token-level prompt compression} \cite{li-etal-2025-prompt}, \textit{i.e.} exploiting the redundancy of human language \cite{Shannon51} and mapping the original sequence to a shorter one, while preserving original semantics.

We focus on the \textit{question-agnostic compression} \cite{jiang-etal-2023-llmlingua}, which aims to process the context provided to the model so that the compression can be re-used for any context-related question or task: possible use-cases include (i) combining compression with Retrieval-Augmented Generation \cite{Gao2023RetrievalAugmentedGF} via retrieving pre-compressed entries, (ii) single-call compressions of lengthy texts such as meeting transcripts \cite{hu-etal-2023-meetingbank} for their subsequent comprehensive analysis involving multiple LLM calls, (iii) optimizing LLM-powered learning platforms via compressing learning materials (iv) addressing tokens-per-minute (TPM) API bandwidth limitations.

The most popular approaches tailored for this set-up are \textbf{extractive} \cite{jiang-etal-2023-llmlingua, pan-etal-2024-llmlingua}, meaning that compression is posed as a binary classification problem of preserving/removing each of the input sequence parts (usually tokens). We hypothesize that \textbf{abstractive} methods, operating under a much larger space of valid compressions, can provide better outputs through paraphrase of the input sequence. To this end, we define $2$ metrics, characterizing the quality of a \textit{Compressor} LLM: (i) \textit{Target} LLM performance on a downstream task given its inputs are preprocessed by the \textit{Compressor} (ii) Adherence of the \textit{Compressor} to the target \textit{Compression Rate (CR)}, reflecting user's tolerance to the quality deterioration versus the incurred costs. We ask the following research questions (RQs):

\vspace{-0.2em}
\begin{itemize}[noitemsep, topsep=0pt]
    \item \textbf{RQ1}. What are the compression capabilities of the off-the-shelf LLMs?
    \item \textbf{RQ2}. Can we further improve their performance with prompt optimization techniques such as   \cite{yuksekgonul2025optimizing}?
    \item \textbf{RQ3} How does performance of an SFT/RL post-trained ``small LLM'' \textit{Compressor} compares to the SOTA \textbf{extractive} approaches across different datasets?
\end{itemize}
\vspace{-0.2em}

\noindent We share results of the extensive benchmarking addressing \textbf{RQ1}, and show that the answer to \textbf{RQ2} is positive in case the prompt is used with the same \textit{Compressor} model it was optimized for. Most importantly, we present \textbf{Cmprsr}, an abstractive \textit{Compressor} outperforming SOTA extractive compression across different CR-s, which answers \textbf{RQ3}.

\section{Related work}
\label{Sec:related work}
 \begin{figure*}[ht]
     \centering
     \includegraphics[width=0.76\linewidth,height=0.6\linewidth]{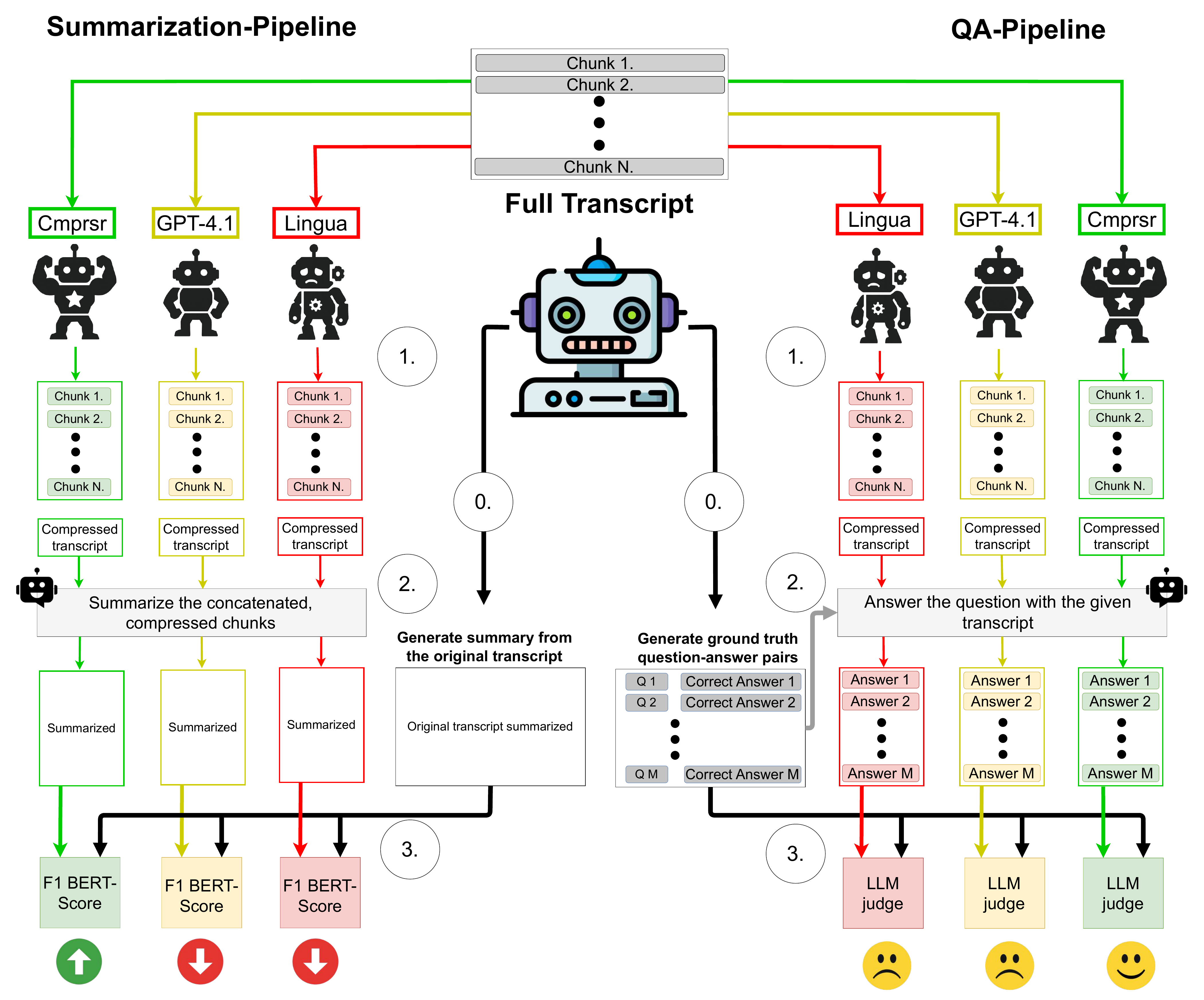}

    \caption{Evaluation pipeline. We assess transcripts' compressions on $2$ downstream tasks. (i) \textbf{Summarization}, i.e we compute BertScore between compressed and original. (ii) \textbf{QA}, where we build a dataset of questions and answers from MeetingBank transcripts, and measure the \textit{Target} model’s accuracy using the compressed context.}

    \label{fig:mb_assessment_pipeline}
 \end{figure*}
 
Compression can be performed in either \textit{question-aware} \cite{shandilya2024tacorltaskawareprompt, kim2025acornnoiserobustabstractivecompression, choi-etal-2024-reading, yoon-etal-2024-compact} or \textit{question-agnostic} way. In the \textit{question-aware} set-up, compression is conditioned on the given question and aims to filter out all irrelevant information. While allowing high CRs, this also prevents compression re-usage for the new queries. Motivated by the use-cases detailed in Sec. \ref{Sec:intro}, we focus on \textit{question-agnostic} set-up.  According to \cite{li-etal-2025-prompt}, prompt compression can also be divided into the following $2$ categories:

\textbf{Embedding-level}  (\textit{a.k.a} soft prompt methods \cite{mu2023learning, chevalier-etal-2023-adapting}), which require access to the \textit{Target} model, and are therefore not viable for the black-box models.

\textbf{Token-level} (\textit{a.k.a} hard prompt methods): (i) \textit{Extractive}: compression space is the set of order-preserving subsequences of the input; the filtering can be implemented on the token \cite{jiang-etal-2023-llmlingua, pan-etal-2024-llmlingua, zhao-etal-2025-dac} or the sentence \cite{Liskavets25} level. Notably, \cite{hu2025dynamiccompressingpromptsefficient, shandilya2024tacorltaskawareprompt} use RL techniques to optimize the extraction policy. Filtering can be informed by information entropy \cite{jiang-etal-2023-llmlingua, li-etal-2023-compressing}, attention scores \cite{zhao-etal-2025-dac} or performed with a classification model \cite{pan-etal-2024-llmlingua} (ii) \textit{Abstractive}: the compression space is all sequences over the vocabulary, which allows semantics-preserving reordering via tokens not present in the input. 

Below, we detail prior contributions falling into the same broad category as ours, \textit{i.e.} \textbf{question-agnostic token-level abstractive compression}. \citet{pu-etal-2024-style} rely on vanilla \textit{LLaMA-2-7B} to perform compression guided by demonstrations, optimized for the particular dataset. Despite improvement over straightforward prompting, this method is not readily generalizable across tasks, as each new dataset requires generating/selecting a new set of demonstrations. In a recent work, \cite{zhang2025scopegenerativeapproachllm} develop another approach to abstractive compression, using either \textit{GPT-4o-mini} or \textit{Qwen-2.5}: they rely on dynamic chunking and chunk-specific CRs, which allows to preserve important context. 

Neither of the two above-mentioned methods compare a meaningful number of LLMs in terms of their aptitude for compression, which, as we show, greatly varies. Furthermore, according to our experiments, even the models dominating the "vanilla compression leaderboard" can be significantly improved with post-training in terms of both compression quality and adherence to the user-specified CR. The work of \cite{chuang-etal-2024-learning} most closely addresses this limitation: while it still experiments with a single backbone (\textit{Vicuna-7B}), the model is actually tuned for better compressions. The main limitation of the presented approach is that it falls short of adopting RL: \textit{Vicuna-7B} serves as both \textit{Compressor} and the \textit{Target} model, and the signal comes from the semantic preservation loss between the original and the compressed input activations. Among the works targeting specific downstream tasks, \citet{larionov-eger-2025-promptoptme} investigate compression for the machine translation quality assessment. While the resulting model cannot be used for prompt compression "in the wild", the authors notably introduce RL (ORPO) for training an abstractive compressor. 

\vspace{-0.1cm}
\section{CompressionBench} \label{CompressionBench}

We focus on the following $2$ metrics reflecting the practical usability of a \textit{Compressor} model: (i) adherence to the desired CR: $CR=n_{tkns}^{cmpr}/n_{tkns}^{original}$; $\Delta_{CR} = CR_{real} - CR_{desired}$, where $desired$ represents the user-defined $CR$, and $real$ stands for the actual one; (ii) performance on the downstream tasks given compressed inputs.

%In case of MeetingBank (MB), $2$ downstream tasks are summarization and question-answering (see Fig. \ref{fig:mb_assessment_pipeline}). The template of the \textit{Compressor} prompt, including the length conditioning---we render CR into the desired number of tokens and add it to the system prompt---is given in the Appendix. Both here and when training \textbf{Cmprsr}, we cut transcripts into chunks before passing them to the \textit{Compressor} model, following the methodology of \cite{pan-etal-2024-llmlingua} to avoid truncated final sentences in the produced chunks. We then combine compressed chunks back into the compressed transcripts. 

Figure~\ref{fig:mb_assessment_pipeline} illustrates the 2 evaluation pipelines used to assess the impact of compression on model utility: \textbf{summarization} and \textbf{question answering (QA)} pipelines. The template of the \textit{Compressor} prompt is provided in the Appendix. It includes length conditioning, where the desired CR is converted into a target token count and appended to the system prompt. For both the benchmarking and the training of \textbf{Cmprsr}, transcripts are segmented into manageable chunks before being passed to the \textit{Compressor} model, following the methodology of \citet{pan-etal-2024-llmlingua} to avoid truncation of incomplete sentences. Compressed chunks are then concatenated to reconstruct the full compressed transcript. We evaluate the resulting compressions through two pipelines:
%\begin{table}[!t]
\begin{table*}[!t]
\renewcommand{\arraystretch}{0.86} % increase row height
\centering
\caption{Compression performance of various vanilla models on the MeetingBank transcripts; truncated version of Table \ref{tab:results_vanilla_mb_detailed}. Within each group, we sort the models based on the average QA performance across compression rates.}
\label{tab:results_vanilla_mb}
\begin{tabular}{l|rrr|rrr|rrr}
\toprule
{} & \multicolumn{3}{c}{$\Delta_{CR}$} & \multicolumn{3}{c}{BERT-F1} & \multicolumn{3}{c}{QA} \\
\cmidrule(lr){2-4}\cmidrule(lr){5-7}\cmidrule(lr){8-10}
Requested CR &                    0.1 &   0.3 &   0.5 &          0.1 &   0.3 &   0.5 &      0.1 &   0.3 &   0.5 \\
% model                              &                        &       &       &              &       &       &          &       &       \\
\midrule
\multicolumn{10}{@{}c}{\textbf{Closed-source models}} \\
\midrule
gpt-5-nano   &                   0.19 &  0.17 &  0.04 &         0.87 &  0.88 &  0.88 &     0.25 &  0.30 &  0.31 \\

\arrayrulecolor{green!30} % ensure border is black
\hline
\rowcolor{green!30} % keep background white (or choose another color)
gpt-4.1-mini & 0.07 & -0.00 & -0.17 & 0.89 & 0.90 & 0.90 & 0.20 & 0.29 & 0.30 \\
\hline
\arrayrulecolor{black} % reset to default for the rest

% gpt-4.1-mini &                   0.07 & 0.00 & -0.17 &         0.89 &  0.90 &  0.90 &     0.20 &  0.29 &  0.30 \\

gpt-5-mini  &                   0.13 &  0.11 &  0.10 &         0.86 &  0.87 &  0.87 &     0.19 &  0.25 &  0.27 \\
gpt-5        &                   0.10 &  0.08 &  0.06 &         0.87 &  0.88 &  0.88 &     0.18 &  0.25 &  0.25 \\
gemini-2.0-flash-lite   &                   0.09 & 0.00 & -0.17 &         0.88 &  0.89 &  0.89 &     0.17 &  0.24 &  0.25 \\
gpt-4.1-nano &                   0.04 & -0.02 & -0.19 &         0.87 &  0.89 &  0.89 &     0.15 &  0.26 &  0.25 \\
gpt-4.1      &                   0.06 & -0.02 & -0.19 &         0.88 &  0.89 &  0.89 &     0.17 &  0.23 &  0.25 \\
gemini-2.5-flash  &                   0.03 & -0.03 & -0.16 &         0.88 &  0.89 &  0.89 &     0.15 &  0.22 &  0.24 \\
o4-mini &                   0.10 &  0.05 & -0.17 &         0.88 &  0.88 &  0.88 &     0.14 &  0.22 &  0.21 \\

\midrule
\multicolumn{10}{@{}c}{\textbf{Large Open-source models ($>10B$)}} \\
\midrule

gemma-3-12b-it                         &                   0.18 &  0.07 & -0.05 &         0.89 &  0.89 &  0.89 &     0.22 &  0.25 &  0.26 \\
Mistral-Small-3.1-24B &                   0.14 & -0.04 & -0.23 &         0.88 &  0.89 &  0.89 &     0.22 &  0.24 &  0.23 \\
DeepSeek-V3                       &                   0.11 & -0.05 & -0.24 &         0.89 &  0.89 &  0.89 &     0.21 &  0.24 &  0.25 \\
% Llama-3.3-70B             &                   0.07 &  0.15 & -0.02 &         0.88 &  0.88 &  0.88 &     0.20 &  0.24 &  0.24 \\
% Qwen3-30B-A3B              &                   0.15 &  0.01 & -0.17 &         0.88 &  0.89 &  0.89 &     0.20 &  0.23 &  0.23 \\
% Qwen3-235B-A22B           &                   0.13 &  0.03 & -0.11 &         0.88 &  0.89 &  0.89 &     0.18 &  0.21 &  0.23 \\
% Qwen2.5-32B                     &                   0.08 & -0.06 & -0.25 &         0.88 &  0.88 &  0.88 &     0.18 &  0.21 &  0.22 \\
% gemma-3-27b                         &                   0.16 &  0.13 &  0.21 &         0.88 &  0.88 &  0.88 &     0.16 &  0.18 &  0.20 \\
% Meta-Llama-3.1-405B       &                   0.04 & -0.10 & -0.30 &         0.86 &  0.87 &  0.87 &     0.18 &  0.17 &  0.18 \\
% Qwen2.5-14B                     &                   0.11 & -0.03 & -0.21 &         0.87 &  0.88 &  0.87 &     0.15 &  0.17 &  0.17 \\

\midrule
\multicolumn{10}{@{}c}{\textbf{Small Open-source models ($<10B$)}} \\
\midrule

Llama-3.2-3B &                   0.05 & -0.10 & -0.30 &         0.87 &  0.87 &  0.87 &     0.17 &  0.21 &  0.22 \\

\arrayrulecolor{green!30} % ensure border is black
\hline
\rowcolor{green!30} % keep background white (or choose another color)
 Qwen3-4B      &                   0.05 & -0.08 & -0.26 &         0.86 &  0.88 &  0.88 &     0.16 &  0.21 &  0.22 \\
\hline
\arrayrulecolor{black} % reset to default for the rest

% Qwen3-4B      &                   0.05 & -0.08 & -0.26 &         0.86 &  0.88 &  0.88 &     0.16 &  0.21 &  0.22 \\
Qwen2.5-7B         &                   0.08 & -0.11 & -0.31 &         0.87 &  0.88 &  0.88 &     0.16 &  0.19 &  0.18 \\
gemma-3-4b            &                  0.00 & -0.15 & -0.35 &         0.87 &  0.87 &  0.88 &     0.13 &  0.17 &  0.19 \\
Llama-3.1-8B &                  -0.02 & -0.19 & -0.39 &         0.85 &  0.86 &  0.86 &     0.14 &  0.17 &  0.15 \\
Qwen2.5-3B         &                   0.09 & -0.08 & -0.24 &         0.85 &  0.86 &  0.86 &     0.10 &  0.12 &  0.12 \\

% \midrule
% \multicolumn{10}{@{}c}{\textbf{Encoder-Decoder Models}} \\
% \midrule

% flan-t5-xxl &                   0.60 &  0.39 &  0.18 &         0.89 &  0.89 &  0.89 &     0.31 &  0.30 &  0.31 \\
% flan-t5-xl  &                   0.16 & -0.04 & -0.24 &         0.87 &  0.87 &  0.87 &     0.17 &  0.17 &  0.16 \\

\midrule
\multicolumn{10}{@{}c}{\textbf{Extractive}} \\
\midrule

\arrayrulecolor{green!30} % ensure border is black
\hline
\rowcolor{green!30} % keep background white (or choose another color)
 llmlingua-2      &                   -0.01 & -0.03 & -0.03 &         0.86 &  0.89 &  0.9 &     0.16 &  0.34 &  0.42 \\
\hline
\arrayrulecolor{black} % reset to default for the rest

% llmlingua-2 &                  -0.01 & -0.03 & -0.03 &         0.86 &  0.89 &  0.9 &     0.16 &  0.34 &  0.42 \\
% llmlingua-1 &                    NaN &   NaN &   NaN &          NaN &   NaN &  NaN &     0.16 &   NaN &   NaN \\

\bottomrule
\end{tabular}
\end{table*}

\textbf{Summarization pipeline.}  The goal of this pipeline is to evaluate how well the compressed transcripts preserve the semantic content necessary for high-quality summaries. We generate summaries for both the original and the compressed transcripts using the same \textit{Target} model, and compute the \textit{BERTScore-F1} between them: a higher BERTScore indicates that the compressed input preserves the key content of the original.

\textbf{QA pipeline.}  This setup assesses whether the compressed context still contains the information needed to answer questions accurately.  We construct a dataset of question–answer pairs derived from MeetingBank transcripts and compare the \textit{Target} model’s accuracy when conditioned on the compressed versus the original contexts.  A smaller performance gap indicates that the \textit{Compressor} successfully preserves task-relevant information.

%In case of MeetingBank (MB), $2$ downstream tasks are summarization and question-answering (see Fig. \ref{fig:mb_assessment_pipeline}). The template of the \textit{Compressor} prompt, including the length conditioning---we render CR into the desired number of tokens and add it to the system prompt---is given in the Appendix. Both here and when training \textbf{Cmprsr}, we cut transcripts into chunks before passing them to the \textit{Compressor} model, following the methodology of \cite{pan-etal-2024-llmlingua} to avoid truncated final sentences in the produced chunks. We then combine compressed chunks back into the compressed transcripts. 

We present the most important MB results in Table \ref{tab:results_vanilla_mb}, and provide full MB results (Table \ref{tab:results_vanilla_mb_detailed}) along with the GSM8k results (Table \ref{tab:results_vanilla_gsm8k}) and the full names of the models (Table \ref{tab:models_name_mapping}) in the Appendix. We dissect the benchmarking results below as a set of enumerated \textbf{findings}: %F1, F2, F3, and F4. 

\begin{enumerate}[leftmargin=2em,label=\textbf{F\arabic*}]
    \item \textbf{Vanilla LLMs poorly adhere to the prompted CR.} While they are susceptible to the prompted rate, the length of the generations skews towards some fixed CR, leading to "under-compression" for high CRs (\textit{0.1}) and "over-compression" for low CRs (\textit{0.5}). \textit{LLMLingua-2} does not suffer from this limitation, as the classification threshold is dynamically adjusted for the extractive methods.  

    \item \textbf{Abstractive LLMs excel at high CRs}. Unlike extractive methods, they can rephrase and condense information beyond the original tokens, preserving key semantics. This highlights the crucial role of abstraction for aggressive compression.

    \item \textbf{Comparison of LLMs.} Although closed-source models generally outperform large open-source models, which in turn surpass smaller open-source ones, performance is not monotonic within each class. Model size or release date alone does not predict the outcome: for example,\textit{ gpt-4.1-mini} outperforms both \textit{gpt-4.1} and \textit{gpt-5-mini} on CR adherence and compression quality.

    \item \textbf{LLMs vastly outperform \textit{LLMLingua-2} on the shorter prompts from GSM8k}, although CR adherence is worse than for long MB prompt (Table \ref{tab:results_vanilla_gsm8k}). 

\end{enumerate}

The results from this section inform our choice of the models for the \textbf{Cmprsr} experiments: \textit{Qwen3-4B} among the small open-source models (it performs on par with \textit{LLama-3.2-3B} with slighly better CR adherence), and \textit{gpt-4.1-mini} among the closed-source ones.

\subsection{Boosting vanilla performance with TextGrad}

%\textbf{Motivation.} The system prompt is crucial for aligning LLM output with a user's expectations. As a result, it has a major impact on the performance of our LLM-based compressor. With the increase in LLM popularity, prompting has grown into a mature field with a variety of techniques, including step-by-step reasoning instructions and few-shot examples. However, choosing \emph{which} technique to use and \emph{how} to phrase it remains nontrivial and highly task-dependent. To solve both of these problems, we use \textsc{TextGrad} \cite{yuksekgonul2024textgradautomaticdifferentiationtext} for principled prompt optimization. \textsc{TextGrad} treats textual components as optimization variables and improves them via natural-language ``gradients.''

\textbf{Motivation.} As LLMs have grown in popularity, prompting has matured, offering diverse techniques such as step-by-step reasoning and few-shot examples. Yet selecting the right technique and phrasing remains nontrivial and task-dependent. To address this, we employ \textsc{TextGrad} \cite{yuksekgonul2024textgradautomaticdifferentiationtext} for principled prompt optimization. \textsc{TextGrad} treats texts as optimization variables, refining them via natural-language “gradients.”

%\textbf{Method.}  We model the MeetingBank QA benchmark as a computation graph with nodes corresponding to stages in the LLM pipeline. Each node is assigned a role description, helping the optimizer-LLM understand the high-level interdependencies among nodes. Starting from the ground-truth–based evaluation at the output, \textsc{TextGrad} (i) identifies failure modes, (ii) generates suggestions on how to fix them, (iii) \emph{backpropagates} this feedback through the graph to the upstream nodes, and (iv) uses the accumulated feedback to update the compressor system prompt. For evaluation, we use a 10-point LLM judge score to assess QA performance, where the LLM is asked to grade each answer and assign a score. Additionally, we evaluate adherence to the target CR. More details on \textsc{TextGrad} are provided in Appendix \ref{textgrad-appendix}.

\textbf{Method.} We model the MeetingBank QA benchmark as a computation graph whose nodes represent LLM pipeline stages. Each node has a role description to help the optimizer-LLM grasp the high-level dependencies. Beginning with ground-truth evaluation at the output, \textsc{TextGrad} (i) identifies failures, (ii) proposes fixes, (iii) \emph{backpropagates} feedback through the graph, and (iv) updates the compressor’s system prompt. We evaluate QA performance using a 10-point LLM judge that grades each answer. We also assess adherence to the target CR. Further details appear in Appendix~\ref{textgrad-appendix}. We applied \textsc{TextGrad} to optimize the system prompt for two well-performing compressor LLMs, gpt-4.1-mini and Qwen3-4B. The optimization trajectories differed significantly between the two models. 

%As \textsc{TextGrad} iteratively updates the prompt, the model outputs become increasingly better aligned with the downstream QA task. Figure \ref{fig:textgrad-judge-qwen} illustrates the relationship between the QA quality ($y$-axis, higher is better) and the adherence to the CR ($x$-axis, higher is better), computed on a 100-transcript hold-out set. The evolution of the optimization trajectory is shown; each point is an iteration of \textsc{TextGrad}. Panels include parts of the initial system prompt (iter.\ 0) and the prompt with the best quality (iter.\ 8).

\begin{figure}[ht]
  \centering
  \includegraphics[width=0.92\linewidth]{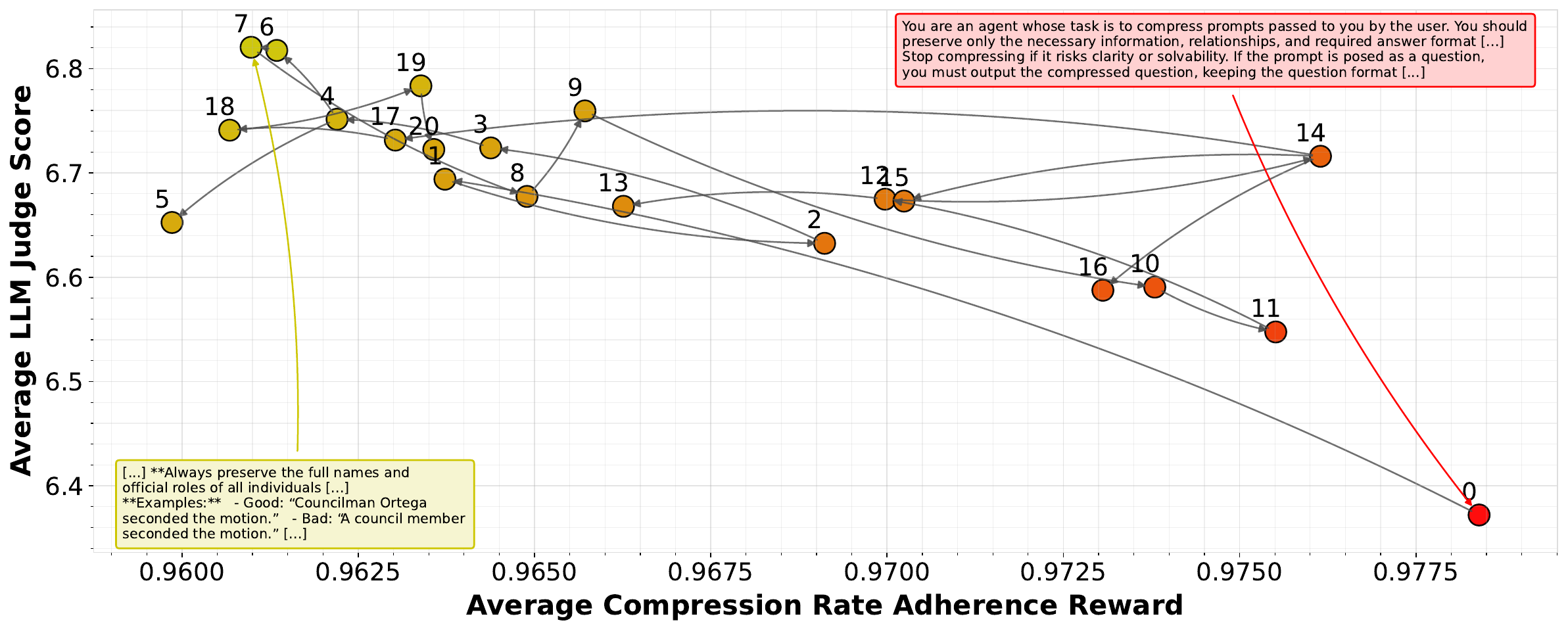}
  \caption{Prompt optimization for Qwen3-4B.}
  \label{fig:textgrad-judge-qwen}
\end{figure}

\textbf{TextGrad for  Qwen3-4B}: Figure~\ref{fig:textgrad-judge-qwen} shows QA quality ($y$-axis) versus compression-rate adherence ($x$-axis) on a 100-transcript hold-out set, with each point representing a \textsc{TextGrad} iteration. Panels display parts of the initial and best prompts.

%The plot illustrates a trade-off: when outputs are allowed to be longer (weaker adherence), they preserve more relevant information, increasing the LLM-Judge score. The optimization trajectory follows an oscillatory pattern in which \textsc{TextGrad} alternates between improving quality (at a small adherence cost) and then tightening adherence (while sacrificing some quality). Across iterations, the learned prompts compose a Pareto front with multiple mutually non-dominated points. This diversity lets us choose a system prompt that matches the application needs, whether prioritizing strict CR adherence or allowing a marginal adherence drop in exchange for a gain in the amount of preserved information.

%The best-quality prompt yields a 0.51 gain in the average LLM Judge score, with only a 0.02 drop in CR adherence. Interestingly, the learned prompt addresses the uncovered failure modes by adding a positive and a negative example as well as by stressing the importance of named entities. As a result, the updated prompt steers the compressor to retain information most salient for answering questions while remaining close to the desired token budget.

The plot reveals a trade-off: longer outputs (weaker adherence) retain more relevant information, boosting the LLM-Judge score. The optimization trajectory oscillates, with \textsc{TextGrad} alternating between improving quality and tightening adherence. The resulting prompts form a Pareto front, enabling choice between strict CR adherence and slightly longer, higher-quality compressions. The best prompt yields a 0.51 gain in the average LLM-Judge score with only a 0.02 drop in CR adherence. 
%It addresses failure modes by adding positive and negative examples and emphasizing named entities, steering the compressor to retain key information while staying within budget.

%In contrast, \textsc{TextGrad} yields more modest quality gains for gpt-4.1-mini. At iteration 33, the optimizer produced a prompt with a strong balance of answer quality and compression length control, resulting in a 0.06 increase in the LLM-Judge score, accompanied by a 0.07 decrease in the CR adherence reward. The learned prompt closely resembles that of Qwen3-4B, and is composed of instructions on what (and how) to retain followed by a set of positive and negative few-shot examples.

\begin{figure}[ht]
  \centering
  \includegraphics[width=0.92\linewidth]{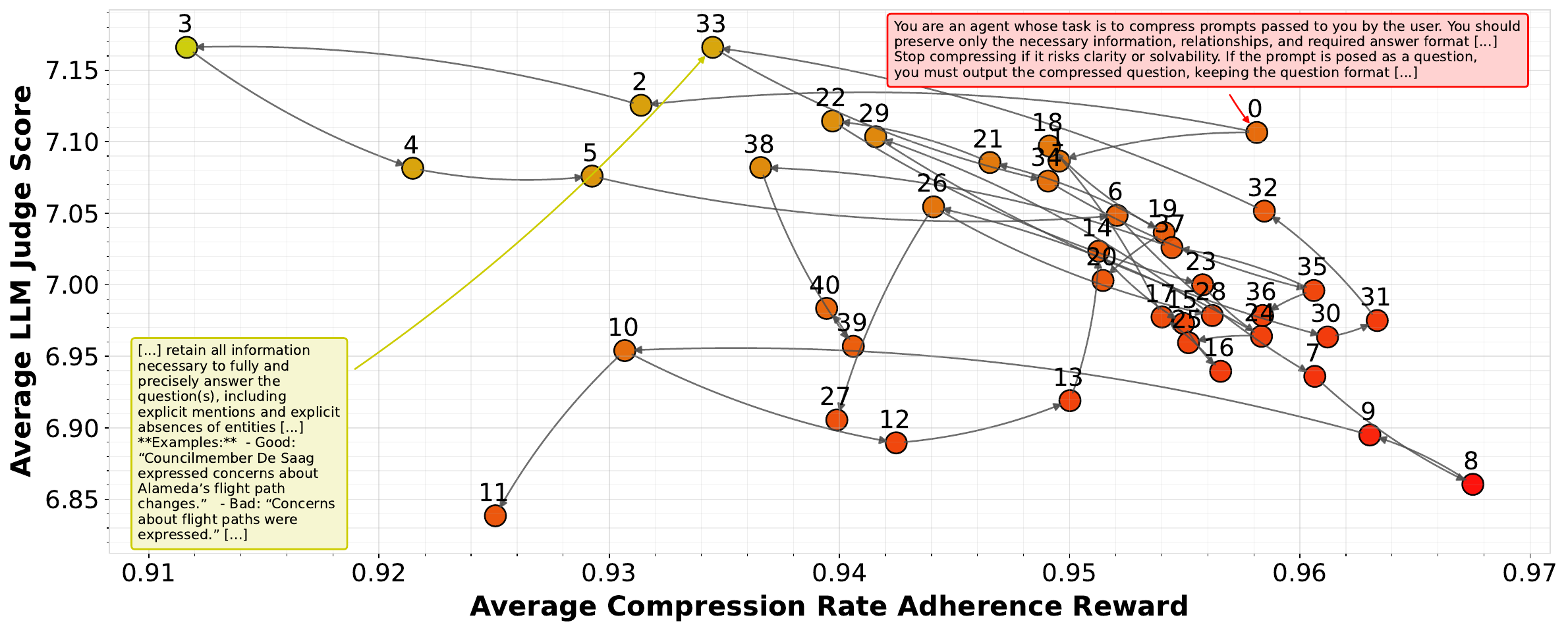}
  \caption{Prompt optimization for gpt-4.1-mini.}
  \label{fig:textgrad-judge-gpt}
\end{figure}

\textbf{TextGrad for gpt-4.1-mini}: In contrast, Figure~\ref{fig:textgrad-judge-gpt} shows that \textsc{TextGrad} produced smaller gains for gpt-4.1-mini. By iteration~33, it achieved a balanced prompt with a 0.06 judge-score increase and a 0.07 drop in adherence. The resulting prompt resembles Qwen3-4B’s, combining retention instructions with positive and negative few-shot examples.

%A plausible explanation of this result is that gpt-4.1-mini (Figure \ref{fig:textgrad-judge-gpt}) starts from a strong baseline performance (judge score 7.11 versus 6.37 for Qwen3-4B), leaving less headroom for \textsc{TextGrad} to uncover frequent and generalizable failure modes. Running for 28 additional iterations (60 total, versus 20 for Qwen3-4B) failed to push the Pareto frontier outward.

This may be because gpt-4.1-mini  starts from a stronger baseline (7.11 vs.\ 6.37 for Qwen3-4B), leaving less room for improvement. Even after 60 iterations (vs.\ 20 for Qwen3-4B), the Pareto frontier did not expand further.

%Notably, for gpt-4.1-mini, the optimizer LLM discovered prompts that substantially improve adherence reward (+0.009) at the cost of reduced judge score (-0.25). These prompts explicitly encourage controlling for the output length (e.g., ``[...] estimate token count and iteratively trim only non-essential content [...].'') This result reflects the optimization trajectory oscillations discussed above. Namely, when the quality improvements level off, \textsc{TextGrad} pivots to the second objective and tightens CR adherence.

Notably, for gpt-4.1-mini, the optimizer found prompts that improved adherence (+0.009) but reduced the judge score (-0.25). These prompts emphasize length control (e.g., “... estimate token count and iteratively trim non-essential content ...”).  This result reflects the optimization trajectory oscillations discussed above.  Namely, when the quality improvements stop, \textsc{TextGrad} pivots to the second objective and tightens CR adherence.

%\textbf{Notes.} Details on our \textsc{TextGrad}-based prompt optimization approach, including a pseudocode implementation, the definition of CR adherence reward, full prompts, and the illustration of the computation graph are provided in Appendix \ref{textgrad-appendix}.

%%%%%%%%%%%%%%%%%%%%%%%%%%

\section{Cmprsr}

% In the previous section, we benchmarked a variety of LLMs on the task of compression, not considering their size. While compression only makes sense if performed by a \textbf{compressor}, which is much smaller than the \textbf{target model}, we are still interested in the "large compressors" as they allow us to collect "gold compressions", which can be used to train the "small compressors". Let us denote the best vanilla compressor overall as \textbf{model T} and the best vanilla compressor among the small models as \textbf{model S}.

% For \textbf{model T}, we 

% \begin{itemize}
%     \item Attempt to further improve performance with Textgrad
%     \item Collect and label compressions: dataset $D$
% \end{itemize}

% For \textbf{model S}, we 

% \begin{itemize}
%     \item Run Knowledge Distillation (KD) on $D$ (simple SFT)
%     \item HIR on $D$: unlike in the previous step, we now make use of the compression labels
%     \item Run GRPO
%     \item Combine the methods above
% \end{itemize}

\subsection{Supervised fine-tuning (SFT)}

%\paragraph{Data generation.} We use the strongest vanilla \textit{Compressor}---\textit{gpt-4.1-mini}---to generate \textit{MB} compressions for the subsequent distillation, while randomly sampling prompted CR from the $[0.1, 0.7]$ range. We notice that the distribution of lengths of the generated data points does not follow the uniform distribution of the prompted rates, skewing to $\approx 0.3$. In order to mitigate that, we experiment with $2$ re-balancing strategies: (i) downsampling, \textit{i.e.,} dropping entries with over-represented lengths, (ii) \textit{upsamplig}, \textit{i.e.,} boosting under-represented ones. We label each generated compression with the number of tokens it contains. 

\paragraph{Data generation.} We use the strongest vanilla \textit{Compressor}, \textit{gpt-4.1-mini}, to generate \textit{MB} compressions for distillation, sampling prompted CRs from $[0.1, 0.7]$. The resulting length distribution deviates from the uniform prompt-rate range, skewing toward $\approx 0.3$. To mitigate this, we test two rebalancing strategies: (i) downsampling, dropping overrepresented lengths, and (ii) \textit{upsampling}, boosting underrepresented ones. Each generated compression is labeled with its token count.

%\paragraph{SFT.} The aim of this stage is twofold: (i) improving \textit{Qwen3-4B} performance through distilling the compression strategies of \textit{gpt-4.1-mini} (ii) improving adherence to the prompted CR. To this end, we append the true number of tokens in the compression to the prompt, i.e., the structure of the sequence the fine-tuning is performed on is \texttt{Original + Length Conditioned Prompt (len(Compression)) + Compression}, and the loss is computed on the \texttt{Compression} part; this ``length conditioning'' is inspired by the Hindsight Instruction Relabelling (\textit{HIR}) works \cite{Zhang23, pie_iclr_2024_spotlight}. In the preliminary experiments, we also tested  additional ``quality conditioning'', \textit{i.e.} passing the normalized quality of the compression relative to other compressions in the same CR range. This research direction, however, had limited success. We varied (\textit{lr}) in the $[10^{-4}, 10^{-8}]$ range, and picked SFT checkpoint with the best validation accuracy, which is the one, trained on the upsampled dataset with $lr=10^{-5}$.

\paragraph{SFT.} This stage has two goals: (i) improving \textit{Qwen3-4B} performance by distilling \textit{gpt-4.1-mini} compression strategies, and (ii) enhancing adherence to the prompted CR. To this end, we append the true token count to the prompt, i.e., the fine-tuning sequence is \texttt{Original + Length Conditioned Prompt (len(Compression)) + Compression}, and the loss is computed on the \texttt{Compression} part; this ``length conditioning'' is inspired by Hindsight Instruction Relabelling (\textit{HIR}) \cite{Zhang23, pie_iclr_2024_spotlight}. Preliminary experiments also tested ``quality conditioning,'' i.e., passing the normalized compression quality relative to others in the same rate range, but it had limited success. We varied the learning rate in $[10^{-4}, 10^{-8}]$ and selected the SFT checkpoint with the best validation accuracy: the one trained on the upsampled dataset with $lr=10^{-5}$.

% We apply supervised fine-tuneing (SFT) to our target model to tune it for adherence to the number of output tokens defined in the user prompt. Each training sample contains the \texttt{original chunk}, and the reference \texttt{compressed\_chunk} that meets the defined expected token limit. The datapairs were extracted from the MeetingBank dataset, and the target compressed chunk was generated by our strongest vanilla compressor (gpt-4.1-mini). The true length distribution of the compressed chunks is skewed toward a CR of $\approx 0.3$, regardless of uniformly sampling between target ratios of $0.1$ and $0.7$. To mitigate this bias, we experimented with two additional re-balancing variants: a \textit{downsampled} variant that flattens over-represented lengths, and an \textit{upsampled} variant that boosts under-represented ratios. We tested our methods on the above-mentioned datasets using several learning rates [$1e-4, ..., 1e-8$] and found the best performing model to be trained on the upsampled dataset using the learning rate of $1e-5$.

\subsection{GRPO}
%To enforce adherence to the target CR and improve compression quality, we extend the SFT stage with GRPO training~\cite{<citation_here>}. Given an original chunk $x$ and its compressed counterpart $x_C$, we define two reward functions.

To ensure that the model not only achieves the desired CR but also retains task-relevant information, we extend the SFT stage with GRPO training. For each input chunk $x$ and its compressed form $x_C$, we define two complementary reward functions: a \textit{length reward} to control compression and a \textit{quality reward} to maintain semantic fidelity.  

\textbf{Length reward.}
The length reward (\ref{length_reward}) encourages the model to adhere to the target CR $r_T$. For a given input, the achieved CR is computed as $r_C = \frac{|x_C|}{|x|}$. This formulation penalizes over-compression (when $r_C > r_T$), while assigning a reward close to $1$ if the compression stays within or below the target threshold. 

\vspace{-3pt}

\begin{equation}
\label{length_reward}
    R_{\text{len}} = 1 - \max\bigl(0, r_C - r_T \bigr),
\end{equation}

\textbf{Quality reward.}  
%The \textbf{quality reward} $R_{\text{qual}}$ (\ref{r_qual}) measures how well the compressed chunk $x_C$ preserves information relevant to the downstream task. We initially experimented with a \textit{QA-based reward}, i.e., the reward is computed by evaluating the compressed text on a QA task against the ground-truth answer from the uncompressed text. However, we found the model exploited this reward by outputting many possible answers instead of compressing effectively. We therefore define a \textit{summary-based reward}, computed as the cross-entropy loss of an open-source solver. Given a query $q$ and answer $a = (a_1, \dots, a_{|a|})$ from chunk $x$, the reward is:

The \textbf{quality reward} $R_{\text{qual}}$ (\ref{r_qual}) measures how well the compressed chunk $x_C$ preserves information relevant to the downstream task. We first experimented with a \textit{QA-based reward}, computed by evaluating the compressed text on a QA task against the ground-truth answer from the uncompressed text. However, the model exploited this reward by outputting many possible answers instead of compressing effectively. %We therefore define a \textit{summary-based reward}, computed as the cross-entropy loss of an open-source solver. 
We therefore define a summary-based reward, where the summary is generated from the original (uncompressed) text. The goal is for the compressed text to produce the same summary as the original when given to the target LLM.Given a query $q$ and answer $a = (a_1, \dots, a_{|a|})$ from chunk $x$, the reward is:

\vspace{-3pt}

\begin{equation}
    R_{\text{qual}} = \frac{1}{1 - \frac{1}{|a|} 
        \sum_{i=1}^{|a|} \log P\bigl(a_i \mid q, x_C, a_{<i}\bigr)},
    \label{r_qual}
\end{equation}
where $|a|$ is the answer length and $a_{<i} = (a_1, \dots, a_{i-1})$ its prefix. Intuitively, the reward increases when $x_C$ enables the solver to predict the answer with high probability, indicating that the compressed input has preserved sufficient task-relevant information.  

\textbf{Dual-objective reward.} 
%We combine the two objectives into a single reward signal. The final reward \ref{final_reward} is defined as the sum of the two components,
To balance compactness and informativeness, we combine both objectives into a single reward signal (Equation \ref{final_reward}). This combined reward simultaneously penalizes the unnecessary verbosity, avoids harmful over compression, and promotes representations that are both efficient and effective for task performance. 
\vspace{-3pt}

\begin{equation}
    R = R_{\text{qual}} + R_{\text{len}}.
    \label{final_reward}
\end{equation}

%encourage the model to produce compressed representation that (i) respect the desired token budget and (ii) retrain the critical information needed for downstream reasoning leading to compressed output that are both efficient and effective for task performance. 

\textbf{Training.} Further details of GRPO training are given in Sec.~\ref{app:grpo-details}. While we also experimented with DPO and ORPO, these approaches did not outperform the GRPO-based pipeline; we describe these setups in Sec.~\ref{sec:supp_dpo_orpo} of the Appendix.

\begin{figure*}[htbp]
    \centering
    \includegraphics[width=\textwidth]{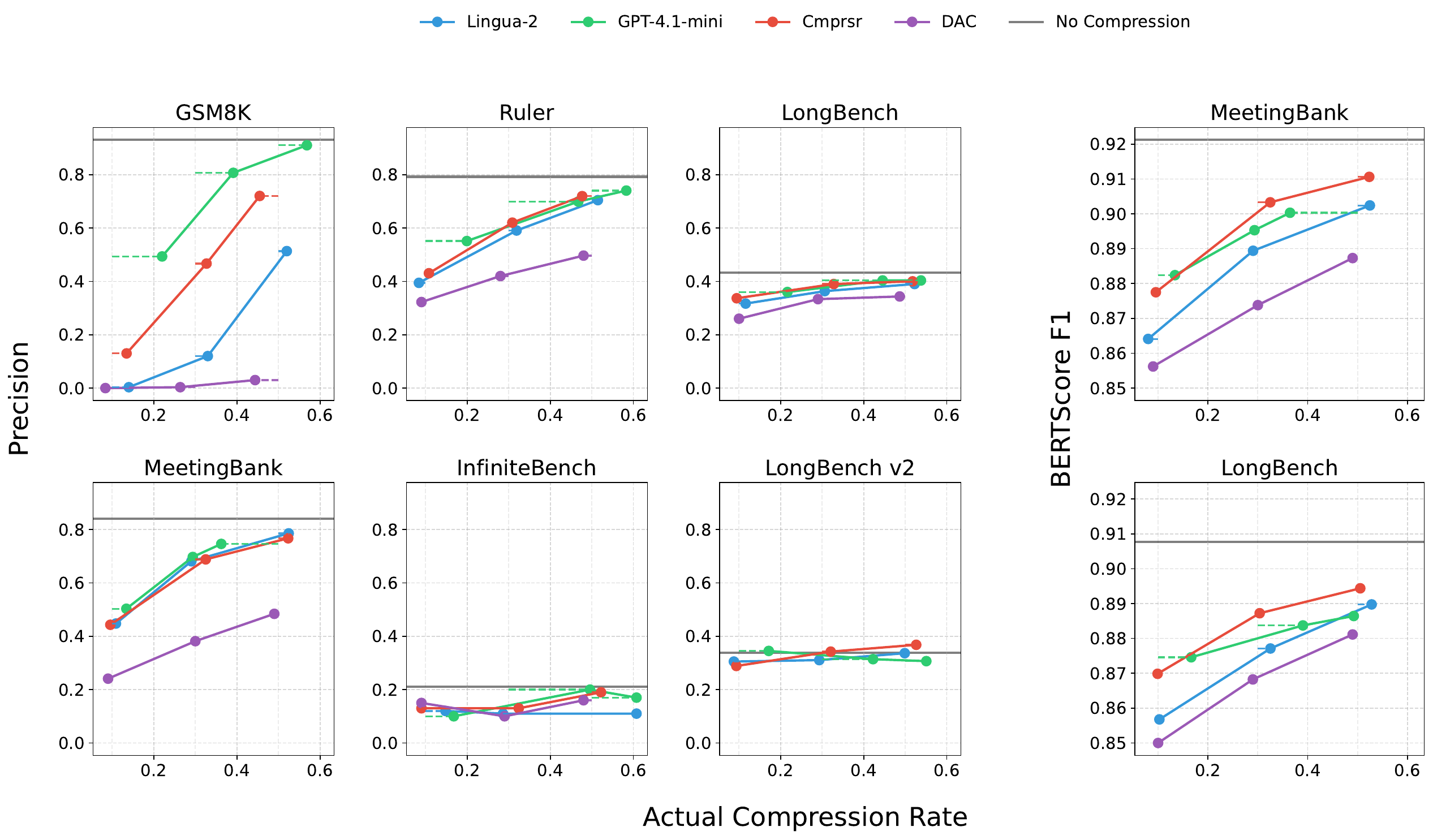}
    \caption{
        Performance comparison across QA and summary benchmarks. 
        Left panel shows precision scores for QA benchmarks (GSM8K, Ruler, LongBench QA, InfiniteBench, LongBench v2, and MeetingBank QA). 
        Right panel shows BERTScore F1 scores for summary benchmarks (MeetingBank Summary and LongBench Summary). 
        The horizontal dashed lines indicate discrepancy between prompted and real compression ratios, and the horizontal black line represents the no-compression baseline. Detailed numerical results for compression rates and performance scores are also provided in Appendix~\ref{A:full_results}. 
    }
    \label{fig:combined_benchmarks}
\end{figure*}
\subsection{Evaluation}

To assess the effectiveness of our approach, we evaluate model performance under two complementary task settings  as explained in Section \ref{CompressionBench}: \textit{summarization} and \textit{QA}. Our primary focus is (i) to examine whether the compressed representations are \textbf{question-agnostic}, i.e., they preserve general information that supports diverse queries, and (ii) to verify whether they retain sufficient \textbf{abstraction} to follow the intent of summarization tasks. Although the model is trained exclusively on the \texttt{MeetingBank} dataset, we evaluate it across multiple out-of-distribution (OOD) benchmarks, to test generalization. These include GSM8K for mathematical reasoning, MeetingBank (QA and Summarization), both version of LongBench, InfiniteBench, and Ruler. The evaluation pipeline operates as follows: a compressed text $x_C$ is first generated, which is then fed into the solver model (\textit{gpt-5-mini}) to produce a summary / answer task-specific questions. 

We compare \textbf{Cmprsr} with $2$ SOTA extractive methods -- \textit{LLMLingua-2} and \textit{DAC} \cite{pan-etal-2024-llmlingua, zhao-etal-2025-dac} -- and vanilla abstractive compression. We drop the original \textit{LLMLingua} \cite{jiang-etal-2023-llmlingua} and \textit{Selective context} \cite{li-etal-2023-compressing}, as they were decisively outperformed by the baselines mentioned above. To achieve a more accurate comparison, we use the same backbone -- \textit{Qwen-3-4B} -- for \textbf{Cmprsr} and \textit{DAC}. 

We report the evaluation results in Figure~\ref{fig:combined_benchmarks} and Table~\ref{tab:full_results} in the Appendix. For summarization tasks, we adopt the Bert-F1 score as the primary metric, while for QA tasks we use precision. For \texttt{LongBench, LongBenchv2, InfiniteBench} and \texttt{Ruler}, which include multiple subtasks within the same dataset, we aggregate results by averaging over QA-related tasks and separately averaging over summarization-related tasks. Full results for each subtask are provided in the Appendix \ref{A:full_results}. Notably, \textit{DAC} performance on \textit{GSM8k} differs significantly from the original paper, as we run it in the zero-shot mode, and compress the question itself, while \cite{zhao-etal-2025-dac} compress demonstrations.

% Since our objective is to maximize task performance under a relaxed constraint of $\Delta_{CR} \leq \epsilon$, we do not simply select the highest-scoring models. Instead, we focus on models that respect the CR budget while still delivering competitive outcomes. For this study, we set $\epsilon$ to 0.03. Note that perfect adherence to the CR target is often unattainable, and a slightly negative $\Delta_{CR}$ is preferable. 

\begin{figure}[ht]
    \centering
    \includegraphics[width=1\linewidth]{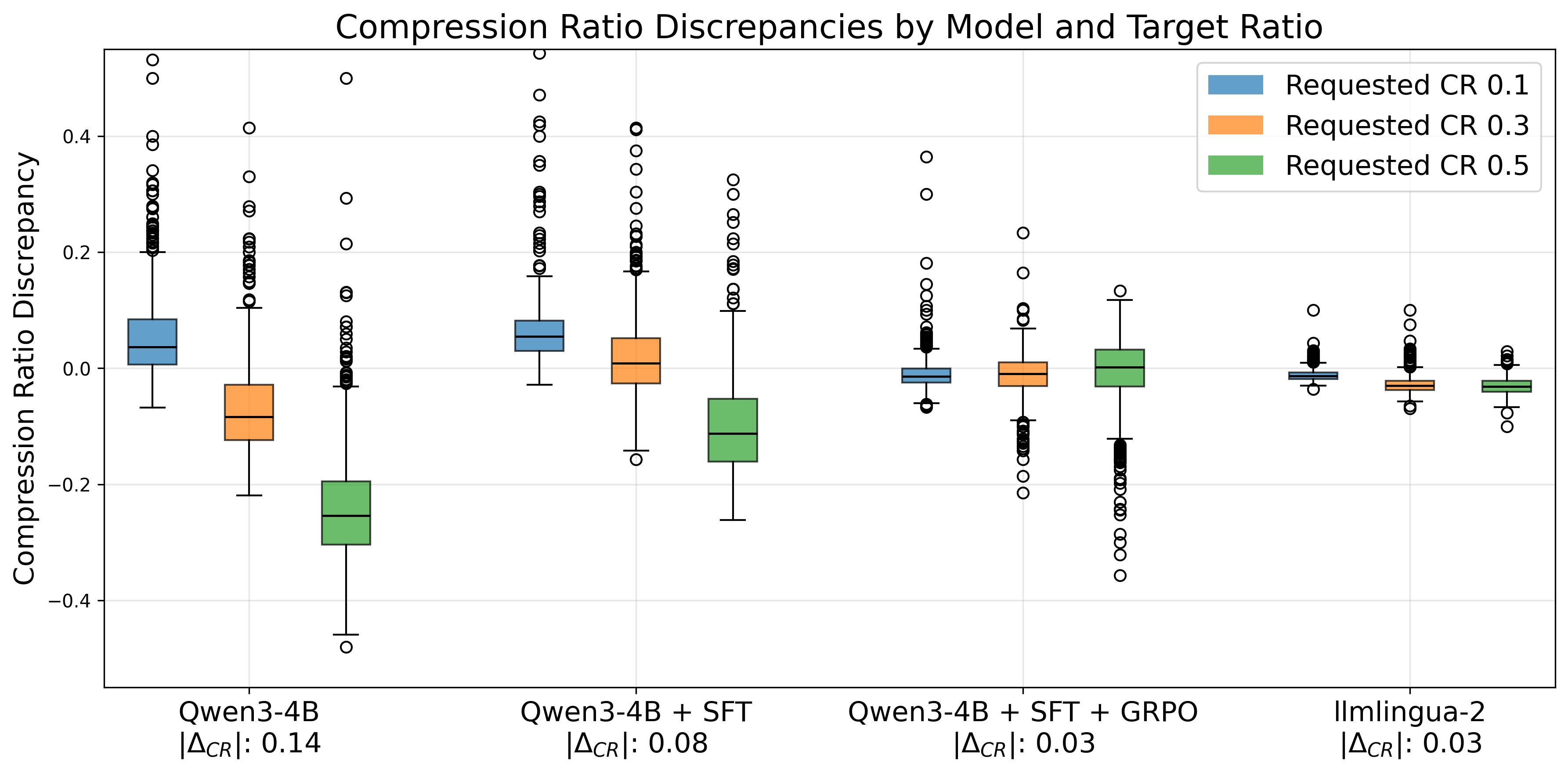}
    \caption{Distribution of $\Delta_{CR}$ across different models, illustrating adherence to the target CR.}
    \label{fig:boxplot}
\end{figure}

Based on Table~\ref{tab:full_results}, higher raw scores are often achieved at the cost of exceeding the target CR, i.e., resulting in positive $\Delta_{CR}$. To account for that, we determine the best method under the constraint of $\Delta_{CR} \leq \epsilon$. Under this criterion, \texttt{Cmprsr} consistently achieves the best trade-off across both QA and summarization tasks. By effectively combining SFT with GRPO, it maintains $\Delta_{CR} \leq 0$, ensuring strict adherence to the target CR while outperforming baseline models. Figure~\ref{fig:boxplot} further illustrates model adherence to the prompted CR across the training stages.

% \texttt{SFT} and \texttt{TextGrad} occasionally yield strong performance, but only when higher $\Delta_{CR}$ values are tolerated, meaning the models compress less than required. 

%To address this, we explore GRPO on top of SFT. This two-stage approach, called \texttt{Cmprsr}, uses a raw-stream version for SFT while applying a distilled version for GRPO. The framework allows us to combine task-specific gradients with compression-aware reinforcement in a second refinement stage. 

\textit{LLMLingua-2} demonstrates stronger adherence due to its inherently extractive compression strategy. In contrast, the Qwen3-4B model fails to match the requested CR, often producing outputs with large negative $\Delta_{CR}$ values. This suggests that, without targeted training, the model tends to over-compress. SFT partially mitigates this issue by reducing the average discrepancy, though it still leaves considerable variance. However, the Cmprsr model, which combines SFT with GRPO, achieves the most balanced results. It keeps discrepancies close to zero while simultaneously reducing variance across all CR. This progression highlights a clear evolution of target adherence throughout the training pipeline: from poor alignment in the vanilla model, to partial improvement with single-objective fine-tuning, to robust adherence with Cmprsr. Moreover, the reduction in variance shows that Cmprsr is not only accurate on average but also reliable at the instance level.

Importantly, Figure \ref{fig:combined_benchmarks} shows that the datasets with the longest entries (\textit{LongBench V2, Infinite bench}) allow for almost lossless compression in terms of the downstream task performance.

\subsection{Analysis}

\begin{figure}[ht]
    \centering
    \includegraphics[width=\linewidth]{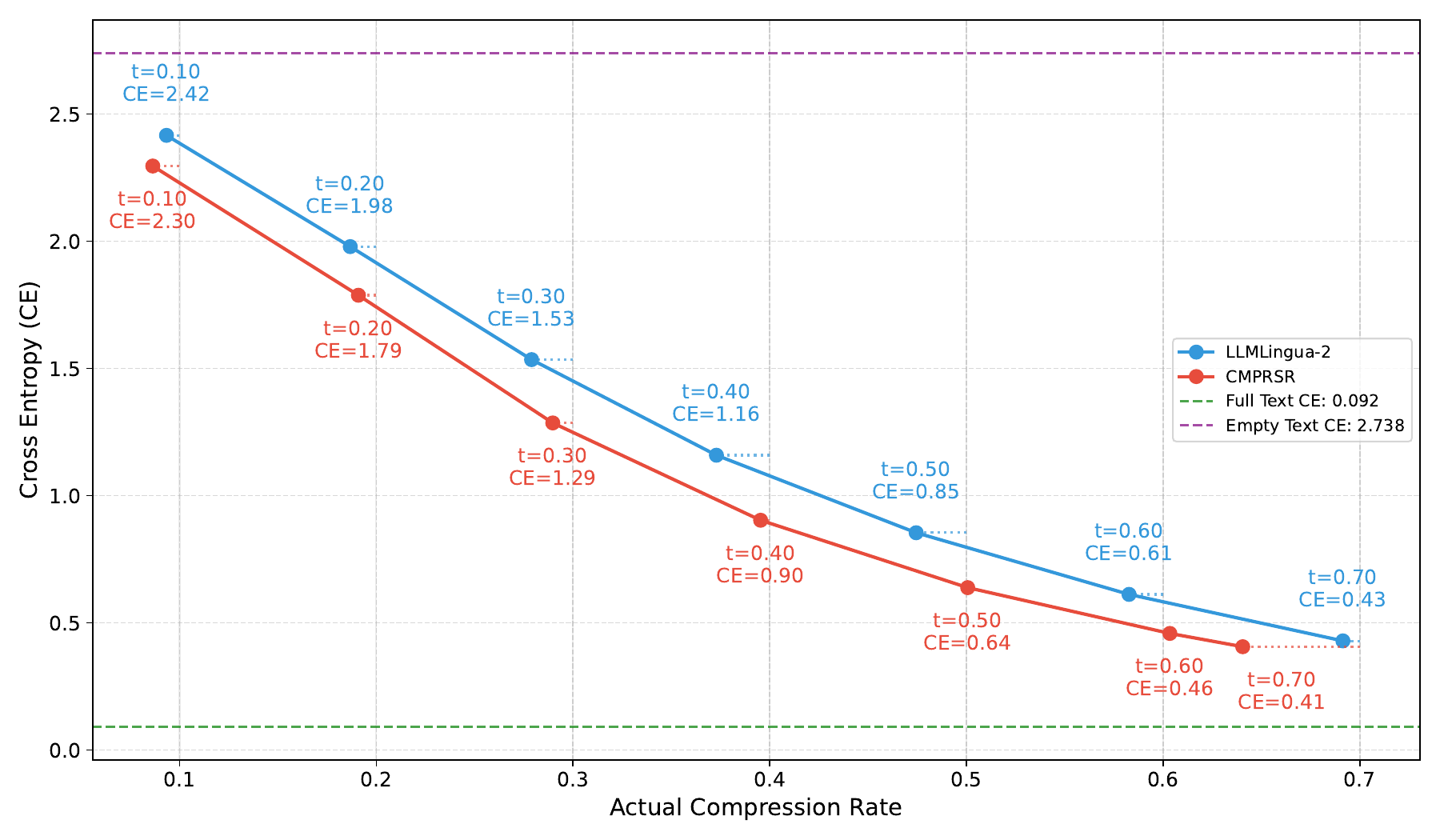}
    \caption{Cross-Entropy (CE) losses of Llama-3.3-70B 4-bit for the summarization task under two prompt compression methods, LLMLingua-2 and Cmprsr, across target CRs. Lower CE indicates better information preservation. The dashed green line marks the mean CE without compression, while the dashed purple line marks CE with empty compression. Numeric tags denote the corresponding target $t$ and $CE$. Results are averaged over 1000 MeetingBank validation samples.}
    \label{fig:ce_llama33}
\end{figure}

\textbf{Cross-entropy under summarization.}
Table \ref{fig:ce_llama33} reports cross-entropy (CE) loss for the summarization task across CRs. \textsc{Cmprsr} yields consistently lower CE than \textsc{llmlingua-2} throughout the range $CR\!\in\![0.1,0.7]$. 
Since both compressors were trained with target rates in this interval, we restrict evaluation accordingly. 
Notably, CE remains near its uncompressed baseline even at $CR{=}0.5$, indicating that substantial prompt reduction can be achieved without materially degrading next-token predictive quality. 
We observe the same trend for \textsc{gemma-3} models of multiple sizes (see Sec. \ref{sec:supp_analysis_gemma} in Appendix), suggesting that \textbf{Cmprsr} generalizes across architectures.

\vspace{-2pt}
\textbf{Local structure preservation.} To probe how compression affects surface form, we compute $n$-gram overlaps between compressed and original prompts (Table \ref{tab:ngram_overlap}). \textbf{Cmprsr} outputs show higher 2-gram and 3-gram overlaps, especially at lower CR. \textbf{Cmprsr} preserves local phrase structure more faithfully, whereas \textsc{llmlingua-2} exhibits higher fragmentation. This pattern supports the hypothesis that \textbf{Cmprsr} maintains semantic adequacy and short-range syntactic cohesion, which can be advantageous for downstream components that rely on multi-token dependencies.

\section{Conclusion}

We propose using LLMs for abstractive prompt compression, starting with a comprehensive benchmark of $25$ off-the-shelf models. Upon that, we improve the best vanilla compressor---\textit{GPT-4.1-mini}---with the \textit{Textgrad}-based meta-prompt optimization, and one of the smaller models---\textit{Qwen-3-4B}---using \textit{SFT} and \textit{GRPO} posttraining. The resulting \textbf{Cmprsr} outperforms both the \textit{extractive} SOTA \textit{LLMLingua-2} and the leading abstractive vanilla -- Textgrad-boosted \textit{gpt-4.1-mini}.

% thereby revealing the full potential of the \textit{abstractive} compression. 

% ---\textit{LLMLingua-2}---across the Pareto frontier. We analyze \textbf{Cmprsr}'s generations, and attribute its performance edge to ...

We argue that "small LLMs" have become capable enough for the practitioners to shift their focus to the paradigm of smaller LLMs compressing prompts for the larger ones. However, vanilla models do not exhibit sufficient performance and should be post-trained for this specific task. \textbf{Cmprsr} both pioneers this line of research and generates SOTA level compressions, suggesting immediate practical value for the community.  

% \newpage

\section{Limitations}

While \textbf{Cmprsr} outperforms the baselines across the Compression Rate spectrum, it is important to note that (i) it is based on a \textbf{4B} backbone, while \textit{e.g.} \textit{LLMLingua-2} is a fine-tuned \textit{xlm-roberta-large} of \textbf{0.55B} parameters (ii) the autoregressive nature of \textbf{Cmprsr} leads to the latency overhead.

(i) implies that, for the compression costs to be offset by savings, \textbf{Cmprsr} should be paired with a sufficiently large \textit{Target} model. (ii) suggests that the prime \textbf{Cmprsr} use-cases are those which allow for the pre-computed compressions: see some of them listed in Sec. \ref{Sec:intro}.

Apart from that, the current version of \textbf{Cmprsr} was trained on a single dataset, suggesting room for improvement and potentially sub-optimal behavior on more specialized OOD data, \textit{e.g.} from coding or medical domain. Further performance gains can also be reached through more elaborate GRPO reward design, as the current reward is biased toward summarization. 

% For example, the inference cost of LLama-3.2-3B -- a model of roughly the same size as \textbf{Cmprsr} -- is $\$0.06$ (both input and output at together.ai) and GPT-5 is $\$1.25$ per $10^6$ input tokens, meaning that in case of $CR=0.3$, the input-incurred GPT-5 cost would be reduced by $64\%$, while in the limit of the free compression the savings would amount to $70\%$. A smaller target model would lead to smaller relative savings though. 

 % Secondly, the usage of either \textbf{0.55B} or a \textbf{4B} model will come at a fraction of the savings, generated by optimized inference of the \textit{Target} model. Quantitatively, LLama-3.2-3B inference cost is $\$0.06$ (both input and output at together.ai) and GPT-5 is $\$1.25$ per $10^6$ input tokens, meaning that in case of $CR=0.3$, the input-incurred GPT-5 cost would be reduced by $64\%$, while in the limit of the free compression the savings would amount to $70\%$.

%\bibliographystyle{acl_natbib}
\bibliography{custom}

@inproceedings{zhao-etal-2025-dac,
    title = "{DAC}: A Dynamic Attention-aware Approach for Task-Agnostic Prompt Compression",
    author = "Zhao, Yi  and
      Li, Zuchao  and
      Zhao, Hai  and
      Qi, Baoyuan  and
      Guoming, Liu",
    editor = "Che, Wanxiang  and
      Nabende, Joyce  and
      Shutova, Ekaterina  and
      Pilehvar, Mohammad Taher",
    booktitle = "Proceedings of the 63rd Annual Meeting of the Association for Computational Linguistics (Volume 1: Long Papers)",
    month = jul,
    year = "2025",
    address = "Vienna, Austria",
    publisher = "Association for Computational Linguistics",
    url = "https://aclanthology.org/2025.acl-long.952/",
    doi = "10.18653/v1/2025.acl-long.952",
    pages = "19395--19407",
    ISBN = "979-8-89176-251-0",
}

@inproceedings{jiang-etal-2023-llmlingua,
    title = "{LLML}ingua: Compressing Prompts for Accelerated Inference of Large Language Models",
    author = "Huiqiang Jiang and Qianhui Wu and Chin-Yew Lin and Yuqing Yang and Lili Qiu",
    editor = "Bouamor, Houda  and
      Pino, Juan  and
      Bali, Kalika",
    booktitle = "Proceedings of the 2023 Conference on Empirical Methods in Natural Language Processing",
    month = dec,
    year = "2023",
    address = "Singapore",
    publisher = "Association for Computational Linguistics",
    url = "https://aclanthology.org/2023.emnlp-main.825",
    doi = "10.18653/v1/2023.emnlp-main.825",
    pages = "13358--13376",
}

@inproceedings{pan-etal-2024-llmlingua,
    title = "{LLML}ingua-2: Data Distillation for Efficient and Faithful Task-Agnostic Prompt Compression",
    author = "Zhuoshi Pan and Qianhui Wu and Huiqiang Jiang and Menglin Xia and Xufang Luo and Jue Zhang and Qingwei Lin and Victor Ruhle and Yuqing Yang and Chin-Yew Lin and H. Vicky Zhao and Lili Qiu and Dongmei Zhang",
    editor = "Ku, Lun-Wei  and
      Martins, Andre  and
      Srikumar, Vivek",
    booktitle = "Findings of the Association for Computational Linguistics ACL 2024",
    month = aug,
    year = "2024",
    address = "Bangkok, Thailand and virtual meeting",
    publisher = "Association for Computational Linguistics",
    url = "https://aclanthology.org/2024.findings-acl.57",
    pages = "963--981",
}

@inproceedings{li-etal-2025-prompt,
    title = "Prompt Compression for Large Language Models: A Survey",
    author = "Li, Zongqian  and
      Liu, Yinhong  and
      Su, Yixuan  and
      Collier, Nigel",
    editor = "Chiruzzo, Luis  and
      Ritter, Alan  and
      Wang, Lu",
    booktitle = "Proceedings of the 2025 Conference of the Nations of the Americas Chapter of the Association for Computational Linguistics: Human Language Technologies (Volume 1: Long Papers)",
    month = apr,
    year = "2025",
    address = "Albuquerque, New Mexico",
    publisher = "Association for Computational Linguistics",
    url = "https://aclanthology.org/2025.naacl-long.368/",
    pages = "7182--7195",
    ISBN = "979-8-89176-189-6",
}

@misc{shandilya2024tacorltaskawareprompt,
      title={TACO-RL: Task Aware Prompt Compression Optimization with Reinforcement Learning}, 
      author={Shivam Shandilya and Menglin Xia and Supriyo Ghosh and Huiqiang Jiang and Jue Zhang and Qianhui Wu and Victor Rühle},
      year={2024},
      eprint={2409.13035},
      archivePrefix={arXiv},
      primaryClass={cs.CL},
      url={https://arxiv.org/abs/2409.13035}, 
}

@inproceedings{pu-etal-2024-style,
    title = "Style-Compress: An {LLM}-Based Prompt Compression Framework Considering Task-Specific Styles",
    author = "Pu, Xiao  and
      He, Tianxing  and
      Wan, Xiaojun",
    editor = "Al-Onaizan, Yaser  and
      Bansal, Mohit  and
      Chen, Yun-Nung",
    booktitle = "Findings of the Association for Computational Linguistics: EMNLP 2024",
    month = nov,
    year = "2024",
    address = "Miami, Florida, USA",
    publisher = "Association for Computational Linguistics",
    url = "https://aclanthology.org/2024.findings-emnlp.851/",
    doi = "10.18653/v1/2024.findings-emnlp.851",
    pages = "14533--14549",
}

@inproceedings{larionov-eger-2025-promptoptme,
    title = "{P}rompt{O}pt{M}e: Error-Aware Prompt Compression for {LLM}-based {MT} Evaluation Metrics",
    author = "Larionov, Daniil  and
      Eger, Steffen",
    editor = "Chiruzzo, Luis  and
      Ritter, Alan  and
      Wang, Lu",
    booktitle = "Proceedings of the 2025 Conference of the Nations of the Americas Chapter of the Association for Computational Linguistics: Human Language Technologies (Volume 1: Long Papers)",
    month = apr,
    year = "2025",
    address = "Albuquerque, New Mexico",
    publisher = "Association for Computational Linguistics",
    url = "https://aclanthology.org/2025.naacl-long.592/",
    doi = "10.18653/v1/2025.naacl-long.592",
    pages = "11807--11820",
    ISBN = "979-8-89176-189-6",
}

@misc{kim2025acornnoiserobustabstractivecompression,
      title={ACoRN: Noise-Robust Abstractive Compression in Retrieval-Augmented Language Models}, 
      author={Singon Kim and Gunho Jung and Seong-Whan Lee},
      year={2025},
      eprint={2504.12673},
      archivePrefix={arXiv},
      primaryClass={cs.CL},
      url={https://arxiv.org/abs/2504.12673}, 
}

@misc{zhang2025scopegenerativeapproachllm,
      title={SCOPE: A Generative Approach for LLM Prompt Compression}, 
      author={Tinghui Zhang and Yifan Wang and Daisy Zhe Wang},
      year={2025},
      eprint={2508.15813},
      archivePrefix={arXiv},
      primaryClass={cs.CL},
      url={https://arxiv.org/abs/2508.15813}, 
}

@inproceedings{choi-etal-2024-reading,
    title = "From Reading to Compressing: Exploring the Multi-document Reader for Prompt Compression",
    author = "Choi, Eunseong  and
      Lee, Sunkyung  and
      Choi, Minjin  and
      Park, Jun  and
      Lee, Jongwuk",
    editor = "Al-Onaizan, Yaser  and
      Bansal, Mohit  and
      Chen, Yun-Nung",
    booktitle = "Findings of the Association for Computational Linguistics: EMNLP 2024",
    month = nov,
    year = "2024",
    address = "Miami, Florida, USA",
    publisher = "Association for Computational Linguistics",
    url = "https://aclanthology.org/2024.findings-emnlp.864/",
    doi = "10.18653/v1/2024.findings-emnlp.864",
    pages = "14734--14754",
}

@inproceedings{Liskavets25,
author = {Liskavets, Barys and Ushakov, Maxim and Roy, Shuvendu and Klibanov, Mark and Etemad, Ali and Luke, Shane K.},
title = {Prompt compression with context-aware sentence encoding for fast and improved LLM inference},
year = {2025},
isbn = {978-1-57735-897-8},
publisher = {AAAI Press},
url = {https://doi.org/10.1609/aaai.v39i23.34639},
doi = {10.1609/aaai.v39i23.34639},
articleno = {2741},
numpages = {10},
series = {AAAI'25/IAAI'25/EAAI'25},
booktitle = ""
}

@misc{hu2025dynamiccompressingpromptsefficient,
      title={Dynamic Compressing Prompts for Efficient Inference of Large Language Models}, 
      author={Jinwu Hu and Wei Zhang and Yufeng Wang and Yu Hu and Bin Xiao and Mingkui Tan and Qing Du},
      year={2025},
      eprint={2504.11004},
      archivePrefix={arXiv},
      primaryClass={cs.CL},
      url={https://arxiv.org/abs/2504.11004}, 
}

@inproceedings{chuang-etal-2024-learning,
    title = "Learning to Compress Prompt in Natural Language Formats",
    author = "Chuang, Yu-Neng  and
      Xing, Tianwei  and
      Chang, Chia-Yuan  and
      Liu, Zirui  and
      Chen, Xun  and
      Hu, Xia",
    editor = "Duh, Kevin  and
      Gomez, Helena  and
      Bethard, Steven",
    booktitle = "Proceedings of the 2024 Conference of the North American Chapter of the Association for Computational Linguistics: Human Language Technologies (Volume 1: Long Papers)",
    month = jun,
    year = "2024",
    address = "Mexico City, Mexico",
    publisher = "Association for Computational Linguistics",
    url = "https://aclanthology.org/2024.naacl-long.429/",
    doi = "10.18653/v1/2024.naacl-long.429",
    pages = "7756--7767",
}

@misc{kaplan2020scalinglawsneurallanguage,
      title={Scaling Laws for Neural Language Models}, 
      author={Jared Kaplan and Sam McCandlish and Tom Henighan and Tom B. Brown and Benjamin Chess and Rewon Child and Scott Gray and Alec Radford and Jeffrey Wu and Dario Amodei},
      year={2020},
      eprint={2001.08361},
      archivePrefix={arXiv},
      primaryClass={cs.LG},
      url={https://arxiv.org/abs/2001.08361}, 
}

@misc{liang2025widespreadadoptionlargelanguage,
      title={The Widespread Adoption of Large Language Model-Assisted Writing Across Society}, 
      author={Weixin Liang and Yaohui Zhang and Mihai Codreanu and Jiayu Wang and Hancheng Cao and James Zou},
      year={2025},
      eprint={2502.09747},
      archivePrefix={arXiv},
      primaryClass={cs.CL},
      url={https://arxiv.org/abs/2502.09747}, 
}

@inproceedings{Luccioni25,
author = {Luccioni, Alexandra Sasha and Strubell, Emma and Crawford, Kate},
title = {From Efficiency Gains to Rebound Effects: The Problem of Jevons' Paradox in AI's Polarized Environmental Debate},
year = {2025},
isbn = {9798400714825},
publisher = {Association for Computing Machinery},
address = {New York, NY, USA},
url = {https://doi.org/10.1145/3715275.3732007},
doi = {10.1145/3715275.3732007},
booktitle = {Proceedings of the 2025 ACM Conference on Fairness, Accountability, and Transparency},
pages = {76–88},
numpages = {13},
location = {
},
series = {FAccT '25}
}

@misc{tully2025midyear
,
  author       = {Tully, Tim and Redfern, Joff and Das, Deedy and Xiao, Derek},
  title        = {2025 Mid-Year LLM Market Update: Foundation Model Landscape + Economics},
  howpublished = {Menlo Ventures Report},
  month        = {July},
  day          = {31},
  year         = {2025},
  url          = {https://menlovc.com/perspective/2025-mid-year-llm-market-update/}
}

@ARTICLE{Shannon51,
  author={Shannon, C. E.},
  journal={The Bell System Technical Journal}, 
  title={Prediction and entropy of printed English}, 
  year={1951},
  volume={30},
  number={1},
  pages={50-64},
  doi={10.1002/j.1538-7305.1951.tb01366.x}}

@article{Gao2023RetrievalAugmentedGF,
  title={Retrieval-Augmented Generation for Large Language Models: A Survey},
  author={Yunfan Gao and Yun Xiong and Xinyu Gao and Kangxiang Jia and Jinliu Pan and Yuxi Bi and Yi Dai and Jiawei Sun and Qianyu Guo and Meng Wang and Haofen Wang},
  journal={ArXiv},
  year={2023},
  volume={abs/2312.10997},
  url={https://api.semanticscholar.org/CorpusID:266359151}
}

@inproceedings{hu-etal-2023-meetingbank,
    title = "MeetingBank: A Benchmark Dataset for Meeting Summarization",
    author = "Yebowen Hu and Tim Ganter and Hanieh Deilamsalehy and Franck Dernoncourt and Hassan Foroosh and Fei Liu",
    booktitle = "Proceedings of the 61st Annual Meeting of the Association for Computational Linguistics (ACL)",
    month = "July",
    year = "2023",
    address = "Toronto, Canada",
    publisher = "Association for Computational Linguistics",
}

@article{yuksekgonul2025optimizing,
  title={Optimizing generative AI by backpropagating language model feedback},
  author={Yuksekgonul, Mert and Bianchi, Federico and Boen, Joseph and Liu, Sheng and Lu, Pan and Huang, Zhi and Guestrin, Carlos and Zou, James},
  journal={Nature},
  volume={639},
  pages={609--616},
  year={2025},
}

@article{mu2023learning,
    title={Learning to Compress Prompts with Gist Tokens}, 
    author={Jesse Mu and Xiang Lisa Li and Noah Goodman},
    year={2023},
    eprint={2304.08467},
    archivePrefix={arXiv},
    primaryClass={cs.CL},
    journal= ""
}

@inproceedings{chevalier-etal-2023-adapting,
    title = "Adapting Language Models to Compress Contexts",
    author = "Chevalier, Alexis  and
      Wettig, Alexander  and
      Ajith, Anirudh  and
      Chen, Danqi",
    editor = "Bouamor, Houda  and
      Pino, Juan  and
      Bali, Kalika",
    booktitle = "Proceedings of the 2023 Conference on Empirical Methods in Natural Language Processing",
    month = dec,
    year = "2023",
    address = "Singapore",
    publisher = "Association for Computational Linguistics",
    url = "https://aclanthology.org/2023.emnlp-main.232/",
    doi = "10.18653/v1/2023.emnlp-main.232",
    pages = "3829--3846",
}

@inproceedings{yoon-etal-2024-compact,
    title = "{C}omp{A}ct: Compressing Retrieved Documents Actively for Question Answering",
    author = "Yoon, Chanwoong  and
      Lee, Taewhoo  and
      Hwang, Hyeon  and
      Jeong, Minbyul  and
      Kang, Jaewoo",
    editor = "Al-Onaizan, Yaser  and
      Bansal, Mohit  and
      Chen, Yun-Nung",
    booktitle = "Proceedings of the 2024 Conference on Empirical Methods in Natural Language Processing",
    month = nov,
    year = "2024",
    address = "Miami, Florida, USA",
    publisher = "Association for Computational Linguistics",
    url = "https://aclanthology.org/2024.emnlp-main.1194/",
    doi = "10.18653/v1/2024.emnlp-main.1194",
    pages = "21424--21439",
}

@misc{bbc2025,
  author       = {BBC},
  title        = {Nasa plans first crewed Moon mission in 50 years for February 2026},
  howpublished = {\url{https://www.bbc.com/news/articles/cy7pegvz17yo}},
  year         = {2025},
  note         = {Accessed: 2025-09-24}
}

@misc{yuksekgonul2024textgradautomaticdifferentiationtext,
      title={TextGrad: Automatic "Differentiation" via Text}, 
      author={Mert Yuksekgonul and Federico Bianchi and Joseph Boen and Sheng Liu and Zhi Huang and Carlos Guestrin and James Zou},
      year={2024},
      eprint={2406.07496},
      archivePrefix={arXiv},
      primaryClass={cs.CL},
      url={https://arxiv.org/abs/2406.07496}, 
}

@inproceedings{Zhang23,
author = {Zhang, Tianjun and Liu, Fangchen and Wong, Justin and Abbeel, Pieter and Gonzalez, Joseph E.},
title = {The wisdom of hindsight makes language models better instruction followers},
year = {2023},
publisher = {JMLR.org},
booktitle = {Proceedings of the 40th International Conference on Machine Learning},
articleno = {1737},
numpages = {15},
location = {Honolulu, Hawaii, USA},
series = {ICML'23}
}

@inproceedings{
pie_iclr_2024_spotlight,
title={Learning Performance-Improving Code Edits},
author={Alexander Shypula and Aman Madaan and Yimeng Zeng and Uri Alon and Jacob R. Gardner and Yiming Yang and Milad Hashemi and Graham Neubig and Parthasarathy Ranganathan and Osbert Bastani and Amir Yazdanbakhsh},
booktitle={The Twelfth International Conference on Learning Representations},
year={2024},
url={https://openreview.net/forum?id=ix7rLVHXyY}
}

@inproceedings{li-etal-2023-compressing,
    title = "Compressing Context to Enhance Inference Efficiency of Large Language Models",
    author = "Li, Yucheng  and
      Dong, Bo  and
      Guerin, Frank  and
      Lin, Chenghua",
    editor = "Bouamor, Houda  and
      Pino, Juan  and
      Bali, Kalika",
    booktitle = "Proceedings of the 2023 Conference on Empirical Methods in Natural Language Processing",
    month = dec,
    year = "2023",
    address = "Singapore",
    publisher = "Association for Computational Linguistics",
    url = "https://aclanthology.org/2023.emnlp-main.391/",
    doi = "10.18653/v1/2023.emnlp-main.391",
    pages = "6342--6353",
}

\appendix

\section{CompressionBench details}
\label{compression_bench}

In this section, we detail the exact versions of the models we use (Table \ref{tab:models_name_mapping}) and the full CompressionBench results on MeetingBank, GSM8k, LongBenchV2 and Ruler.

\subsection{Initial Compressor System Prompt. } The following prompt was used as the initial system prompt in iteration 0 of \textsc{TextGrad}. Additionally, it was used in most experiments and benchmarking across GSM8k, LongBench, and MeetingBank, unless stated otherwise. This prompt was explicitly chosen to be dataset-agnostic and work with most downstream applications.

\begin{tcolorbox}[promptbox]
You are an agent whose task is to compress prompts passed to you by the user. Preserve only the necessary information, relationships, and required answer format. Remove all unnecessary details, drop redundant information, and rephrase what can be rephrased without information loss. You can aggressively shorten words, drop prepositions and articles - do whatever it takes to shorten the prompt to the absolute bare minimum, while avoiding the loss of \textbf{any} important information contained in the prompt. Use compact notation, single-letter variables, and standard abbreviations. Clarify ambiguities minimally. Stop compressing if it risks clarity or solvability. If the prompt is posed as a question, you must output the compressed question, keeping the question format. Do not answer the question; only compress it.
\end{tcolorbox}

\subsection{Models}
We present all models using their short names, while ensuring that the corresponding full model names are introduced when first mentioned.

\begin{table}[ht]
\centering
\caption{Models and their short names}
\label{tab:models_name_mapping}
\resizebox{\columnwidth}{!}{% This forces the table to fit the single column width
\begin{tabular}{l|l}
\toprule
\textbf{Full name} & \textbf{Short name} \\
\midrule
gpt-5-nano-2025-08-07 & gpt-5-nano \\
gpt-4.1-mini-2025-04-14 & gpt-4.1-mini \\
gpt-5-mini-2025-08-07 & gpt-5-mini \\
gpt-5-2025-08-07 & gpt-5 \\
gemini-2.0-flash-lite & gemini-2.0-flash-lite \\
gpt-4.1-nano-2025-04-14 & gpt-4.1-nano \\
gpt-4.1-2025-04-14 & gpt-4.1 \\
gemini-2.5-flash  & gemini-2.5-flash \\
o4-mini-2025-04-16 & o4-mini-2025-04-16 \\
\midrule
gemma-3-12b-it & gemma-3-12b-it \\
Mistral-Small-3.1-24B-Instruct-2503 & Mistral-Small-3.1-24B \\
DeepSeek-V3 & DeepSeek-V3 \\
Llama-3.3-70B-Instruct & Llama-3.3-70B \\
Qwen3-30B-A3B-Instruct-2507 & Qwen3-30B-A3B \\
Qwen3-235B-A22B-Instruct-2507 & Qwen3-235B-A22B \\
Qwen2.5-32B-Instruct & Qwen2.5-32B \\
gemma-3-27b-it & gemma-3-27b \\
Meta-Llama-3.1-405B-Instruct & Meta-Llama-3.1-405B \\
Qwen2.5-14B-Instruct & Qwen2.5-14B \\
\midrule
Llama-3.2-3B-Instruct & Llama-3.2-3B \\
Qwen3-4B-Instruct-2507 & Qwen3-4B \\
Qwen2.5-7B-Instruct & Qwen2.5-7B \\
gemma-3-4b-it & gemma-3-4b \\
Llama-3.1-8B-Instruct & Llama-3.1-8B \\
Qwen2.5-3B-Instruct & Qwen2.5-3B \\
\midrule
flan-t5-xxl & flan-t5-xxl \\
flan-t5-xl & flan-t5-xl \\
\bottomrule
\end{tabular}%
}
\end{table}

\subsection{CompressionBench results on MeetingBank}

We present the full results for all models used to benchmark the LLM as a compressor. Our evaluation includes 9 closed-source models and 18 open-source models spanning a wide range of parameter scales and architectural families. All experiments were conducted on the validation split of the MeetingBank dataset, which provides diverse and realistic meeting transcripts for evaluation. From this split, we randomly selected 100 samples, each of which was divided into multiple chunks when exceeding the 512-token limit (the exact number of chunks varies depending on the tokenizer used).

\subsection{CompressionBench results on GSM8k}

We also report results using the GSM8k dataset to further evaluate the models under a different task setting. In this case, all input sequences are shorter than 512 tokens, so no chunking was required during preprocessing. We randomly selected 300 samples from the validation split of the dataset, ensuring sufficient coverage to assess the performance of both open- and closed-source models under this setting. In the math dataset, where the context is very short, LLM-Lingua performs poorly, which is expected since extractive models cannot effectively compress mathematical problem statements without abstraction, rephrasing, or reordering of words. Meanwhile, the decoder-only vanilla model shows relatively strong performance but does not adhere to the target CR.

\begin{table}[ht]
\centering
\caption{Compression performance of various vanilla models on the GSM8k problems.}
\renewcommand{\arraystretch}{0.85} % increase row height
\label{tab:results_vanilla_gsm8k}
\resizebox{\columnwidth}{!}{% This forces the table to fit the single column width

\begin{tabular}{l|rrr|rrr}
\toprule
{} & \multicolumn{3}{c}{$\Delta_{CR}$} & \multicolumn{3}{c}{Accuracy} \\
\cmidrule(lr){2-4}\cmidrule(lr){5-7}
Requested CR & 0.1 & 0.3 & 0.5 & 0.1 & 0.3 & 0.5 \\
\midrule
\multicolumn{7}{@{}c}{\textbf{Closed-source models}} \\
\midrule
gpt-5-nano   & 0.41 & 0.44 & 0.25 & 0.69 & 0.82 & 0.87 \\
gpt-5        & 0.23 & 0.22 & 0.13 & 0.61 & 0.86 & 0.90 \\
gpt-5-mini   & 0.20 & 0.23 & 0.12 & 0.55 & 0.85 & 0.88 \\
gpt-4.1      & 0.22 & 0.17 & 0.05 & 0.58 & 0.82 & 0.86 \\
gemini-2.5-flash  & 1.27 & 0.17 & 0.17 & 0.52 & 0.79 & 0.87 \\
gemini-2.0-flash-lite   & 0.34 & 0.31 & 0.16 & 0.55 & 0.78 & 0.80 \\
gpt-4.1-mini & 0.20 & 0.17 & 0.06 & 0.43 & 0.80 & 0.85 \\
gpt-4.1-nano-2025-04-14 & 0.10 & 0.13 & 0.01 & 0.17 & 0.62 & 0.76 \\
\midrule
\multicolumn{7}{@{}c}{\textbf{Large Open-source models ($>10$B)}} \\
\midrule
Meta-Llama-3.1-405B & 0.40 & 0.33 & 0.13 & 0.78 & 0.84 & 0.84 \\
Qwen3-235B-A22B     & 0.36 & 0.21 & 0.04 & 0.77 & 0.81 & 0.84 \\
Qwen2.5-32B         & 0.40 & 0.24 & 0.06 & 0.77 & 0.80 & 0.83 \\
DeepSeek-V3          & 0.34 & 0.21 & 0.04 & 0.70 & 0.80 & 0.83 \\
Qwen3-30B-A3B        & 0.36 & 0.21 & 0.04 & 0.69 & 0.77 & 0.80 \\
gemma-3-27b-it       & 0.33 & 0.32 & 0.45 & 0.66 & 0.74 & 0.80 \\
Mistral-Small-3.1-24B & 0.41 & 0.28 & 0.10 & 0.67 & 0.75 & 0.77 \\
Llama-3.3-70B        & 0.25 & 0.22 & 0.06 & 0.51 & 0.73 & 0.81 \\
gemma-3-12b-it       & 0.29 & 0.28 & 0.13 & 0.45 & 0.74 & 0.74 \\
Qwen2.5-14B          & 0.29 & 0.20 & 0.02 & 0.49 & 0.70 & 0.70 \\
\midrule
\multicolumn{7}{@{}c}{\textbf{Small Open-source models ($<10$B)}} \\
\midrule
Qwen3-4B-Instruct & 0.37 & 0.22 & 0.05 & 0.70 & 0.74 & 0.78 \\
Llama-3.1-8B      & 0.36 & 0.23 & 0.05 & 0.55 & 0.62 & 0.63 \\
Qwen2.5-7B        & 0.27 & 0.14 & -0.04 & 0.50 & 0.57 & 0.64 \\
gemma-3-4b         & 0.15 & 0.13 & -0.01 & 0.15 & 0.45 & 0.58 \\
Llama-3.2-3B      & 0.25 & 0.19 & 0.02 & 0.24 & 0.37 & 0.39 \\
Qwen2.5-3B        & 0.08 & 0.12 & -0.11 & 0.15 & 0.21 & 0.28 \\
\midrule
\multicolumn{7}{@{}c}{\textbf{Encoder-Decoder Models}} \\
\midrule
llmlingua-2 & 0.02 & 0.02 & -0.00 & 0.01 & 0.19 & 0.49 \\
\bottomrule
\end{tabular}
}
\end{table}

\begin{table*}[ht]
\centering
\renewcommand{\arraystretch}{0.85} % increase row height
\caption{Compression performance of various vanilla models on the MeetingBank transcripts.}
\label{tab:results_vanilla_mb_detailed}
\resizebox{0.85\textwidth}{0.33\textheight}{%
\begin{tabular}{l|rrr|rrr|rrr}
\toprule
\multirow{2}{*}{ratio} & \multicolumn{3}{c}{$\Delta_{CR}$} & \multicolumn{3}{c}{BERT-F1} & \multicolumn{3}{c}{QA} \\
\cmidrule(lr){2-4}\cmidrule(lr){5-7}\cmidrule(lr){8-10}
 &                    0.1 &   0.3 &   0.5 &          0.1 &   0.3 &   0.5 &      0.1 &   0.3 &   0.5 \\
% model                              &                        &       &       &              &       &       &          &       &       \\
\midrule
\multicolumn{10}{@{}c}{\textbf{Closed-source models}} \\
\midrule
gpt-5-nano   &                   0.19 &  0.17 &  0.04 &         0.87 &  0.88 &  0.88 &     0.25 &  0.30 &  0.31 \\

\arrayrulecolor{green!30} % ensure border is black
\hline
\rowcolor{green!30} % keep background white (or choose another color)
gpt-4.1-mini & 0.07 & -0.00 & -0.17 & 0.89 & 0.90 & 0.90 & 0.20 & 0.29 & 0.30 \\
\hline
\arrayrulecolor{black} % reset to default for the rest

% gpt-4.1-mini &                   0.07 & 0.00 & -0.17 &         0.89 &  0.90 &  0.90 &     0.20 &  0.29 &  0.30 \\

gpt-5-mini  &                   0.13 &  0.11 &  0.10 &         0.86 &  0.87 &  0.87 &     0.19 &  0.25 &  0.27 \\
gpt-5        &                   0.10 &  0.08 &  0.06 &         0.87 &  0.88 &  0.88 &     0.18 &  0.25 &  0.25 \\
gemini-2.0-flash-lite   &                   0.09 & 0.00 & -0.17 &         0.88 &  0.89 &  0.89 &     0.17 &  0.24 &  0.25 \\
gpt-4.1-nano &                   0.04 & -0.02 & -0.19 &         0.87 &  0.89 &  0.89 &     0.15 &  0.26 &  0.25 \\
gpt-4.1      &                   0.06 & -0.02 & -0.19 &         0.88 &  0.89 &  0.89 &     0.17 &  0.23 &  0.25 \\
gemini-2.5-flash  &                   0.03 & -0.03 & -0.16 &         0.88 &  0.89 &  0.89 &     0.15 &  0.22 &  0.24 \\
o4-mini &                   0.10 &  0.05 & -0.17 &         0.88 &  0.88 &  0.88 &     0.14 &  0.22 &  0.21 \\

\midrule
\multicolumn{10}{@{}c}{\textbf{Large Open-source models ($>10B$)}} \\
\midrule

gemma-3-12b-it                         &                   0.18 &  0.07 & -0.05 &         0.89 &  0.89 &  0.89 &     0.22 &  0.25 &  0.26 \\
Mistral-Small-3.1-24B &                   0.14 & -0.04 & -0.23 &         0.88 &  0.89 &  0.89 &     0.22 &  0.24 &  0.23 \\
DeepSeek-V3                       &                   0.11 & -0.05 & -0.24 &         0.89 &  0.89 &  0.89 &     0.21 &  0.24 &  0.25 \\
Llama-3.3-70B             &                   0.07 &  0.15 & -0.02 &         0.88 &  0.88 &  0.88 &     0.20 &  0.24 &  0.24 \\
Qwen3-30B-A3B              &                   0.15 &  0.01 & -0.17 &         0.88 &  0.89 &  0.89 &     0.20 &  0.23 &  0.23 \\
Qwen3-235B-A22B           &                   0.13 &  0.03 & -0.11 &         0.88 &  0.89 &  0.89 &     0.18 &  0.21 &  0.23 \\
Qwen2.5-32B                     &                   0.08 & -0.06 & -0.25 &         0.88 &  0.88 &  0.88 &     0.18 &  0.21 &  0.22 \\
gemma-3-27b                         &                   0.16 &  0.13 &  0.21 &         0.88 &  0.88 &  0.88 &     0.16 &  0.18 &  0.20 \\
Meta-Llama-3.1-405B       &                   0.04 & -0.10 & -0.30 &         0.86 &  0.87 &  0.87 &     0.18 &  0.17 &  0.18 \\
Qwen2.5-14B                     &                   0.11 & -0.03 & -0.21 &         0.87 &  0.88 &  0.87 &     0.15 &  0.17 &  0.17 \\
\midrule
\multicolumn{10}{@{}c}{\textbf{Small Open-source models ($<10B$)}} \\
\midrule

Llama-3.2-3B &                   0.05 & -0.10 & -0.30 &         0.87 &  0.87 &  0.87 &     0.17 &  0.21 &  0.22 \\

\arrayrulecolor{green!30} % ensure border is black
\hline
\rowcolor{green!30} % keep background white (or choose another color)
 Qwen3-4B      &                   0.05 & -0.08 & -0.26 &         0.86 &  0.88 &  0.88 &     0.16 &  0.21 &  0.22 \\
\hline
\arrayrulecolor{black} % reset to default for the rest

% Qwen3-4B      &                   0.05 & -0.08 & -0.26 &         0.86 &  0.88 &  0.88 &     0.16 &  0.21 &  0.22 \\
Qwen2.5-7B         &                   0.08 & -0.11 & -0.31 &         0.87 &  0.88 &  0.88 &     0.16 &  0.19 &  0.18 \\
gemma-3-4b            &                  0.00 & -0.15 & -0.35 &         0.87 &  0.87 &  0.88 &     0.13 &  0.17 &  0.19 \\
Llama-3.1-8B &                  -0.02 & -0.19 & -0.39 &         0.85 &  0.86 &  0.86 &     0.14 &  0.17 &  0.15 \\
Qwen2.5-3B         &                   0.09 & -0.08 & -0.24 &         0.85 &  0.86 &  0.86 &     0.10 &  0.12 &  0.12 \\

\midrule
\multicolumn{10}{@{}c}{\textbf{Encoder-Decoder Models}} \\
\midrule

flan-t5-xxl &                   0.60 &  0.39 &  0.18 &         0.89 &  0.89 &  0.89 &     0.31 &  0.30 &  0.31 \\
flan-t5-xl  &                   0.16 & -0.04 & -0.24 &         0.87 &  0.87 &  0.87 &     0.17 &  0.17 &  0.16 \\

\midrule
\multicolumn{10}{@{}c}{\textbf{Extractive}} \\
\midrule

\rowcolor{green!30}
llmlingua-2 &                  -0.01 & -0.03 & -0.03 &         0.86 &  0.89 &  0.9 &     0.16 &  0.34 &  0.42 \\
%llmlingua-1 &                    NaN &   NaN &   NaN &          NaN &   NaN &  NaN &     0.16 &   NaN &   NaN \\

\bottomrule
\end{tabular}
}
\end{table*}

\section{\textbf{Cmprsr} details}
\subsection{GRPO Training Details}
\label{app:grpo-details}
We fine-tune the compressor with Grouped Relative Preference Optimization (GRPO) on a $10$k-example subset of the SFT training split. 
For each input, target CR is uniformly distributed with $r \sim \mathcal{U}(0.1, 0.7)$.

Training runs on two NVIDIA A100 GPUs: one GPU performs optimization, and the other handles rollout generation and CE computation, decoupling sampling from updates. We use the frozen Qwen3-4B Instruct model for the CE computations. All the hype parameters are detailed in Table \ref{tab:hyperparameters}:

%\paragraph{Key hyperparameters.}
% We train for two epochs with AdamW (8-bit) and a linear schedule with warmup. Mixed precision uses \texttt{bf16}. 
%Per-device batch size is $8$ with gradient accumulation of $16$.
\begin{table}[ht]
\centering
\renewcommand{\arraystretch}{0.85} % Increased to 1.2 for better readability
\caption{Training Hyperparameters}
\label{tab:hyperparameters}

\resizebox{0.45\textwidth}{!}{% % Resize box now wraps ONLY the tabular part
\begin{tabular}{ll}
\toprule
Setting & Value \\
\midrule
Rollouts per input & $G=4$ \\
Epochs & $2$ \\
Optimizer & AdamW (8-bit) \\
Learning rate & $5\times10^{-6}$ \\
Weight decay & $0.01$ \\
Scheduler / Warmup & Linear / $10\%$ \\
Per-device train batch & $8$ \\
Gradient accumulation & $16$ \\
Precision & \texttt{bf16} \\
\bottomrule
\end{tabular}
} % End resize box
\end{table}

\subsection{DPO and ORPO}
\label{sec:supp_dpo_orpo}

We apply DPO to learn compression policies under a fixed token budget. Each training sample contains a prompt (with the conditions seen before) with the original chunk (extracted from MeetingBank) and a pair of candidate compressions: a \emph{chosen} output produced by a stronger, teacher model (gpt-4.1-mini) and a rejected output produced by a weaker, student model (\texttt{gpt-4.1-nano}). To make sure that for each chunk the teacher-produced one contains higher quality compression, we calculate the \textbf{BERT-F1} score between the compressed and the original chunks and we select the triplets only if the teacher has produced higher F1-scored compressions. We evaluate DPO under two initializations (i) from the vanilla model and (ii) from the SFT-initialized model. We also evaluate ORPO, a reference-free variant that simplifies preference optimization to a single-model setup while retaining the same paired (chosen/rejected) data  using the same setup to enable like-for-like comparisons

\subsection{Training Data Generation}
\begin{figure}[ht]
    \centering
    \includegraphics[width=1\linewidth]{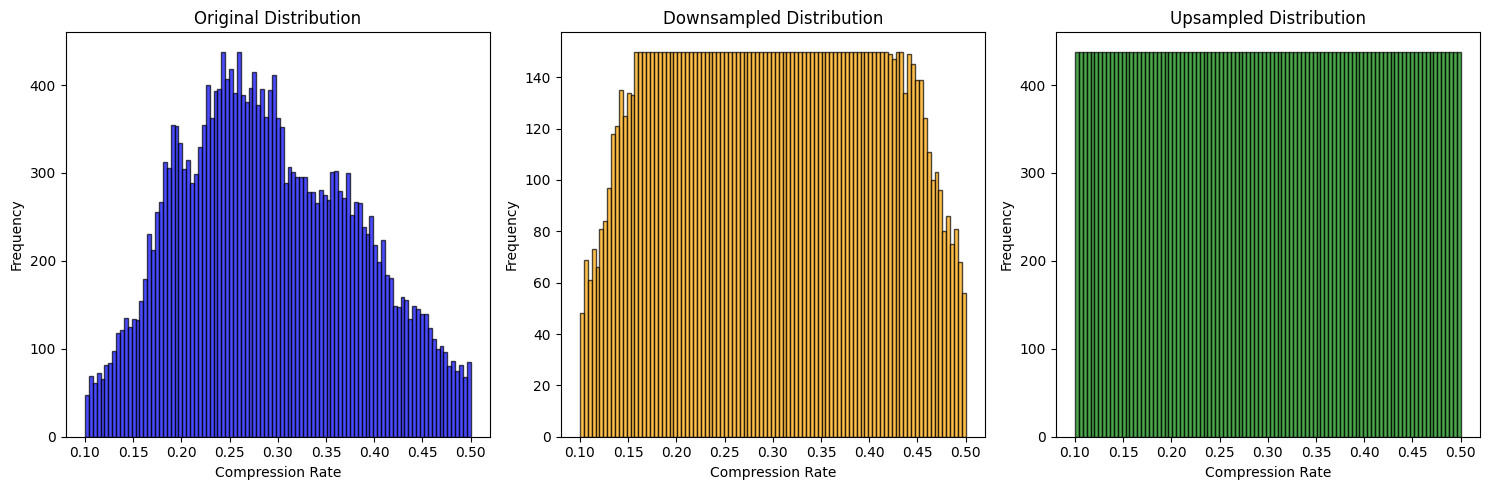}
    \caption{Data distribution}
    \label{fig:data_dist}
\end{figure}

Figure \ref{fig:data_dist} shows the data distribution in the dataset generated to train the SFT backbone. The blue plot shows the original distribution which is favouring ~$0.3$ CR. To mitigate this, two techniques are applied: yellow showcases the downsampled version of the dataset, where we applied a cut at 150 samples in each bin, while the green showcases the oversampled variant.

\subsection{Plots}
\begin{figure}[!htbp]
    \centering
    \includegraphics[width=\linewidth]{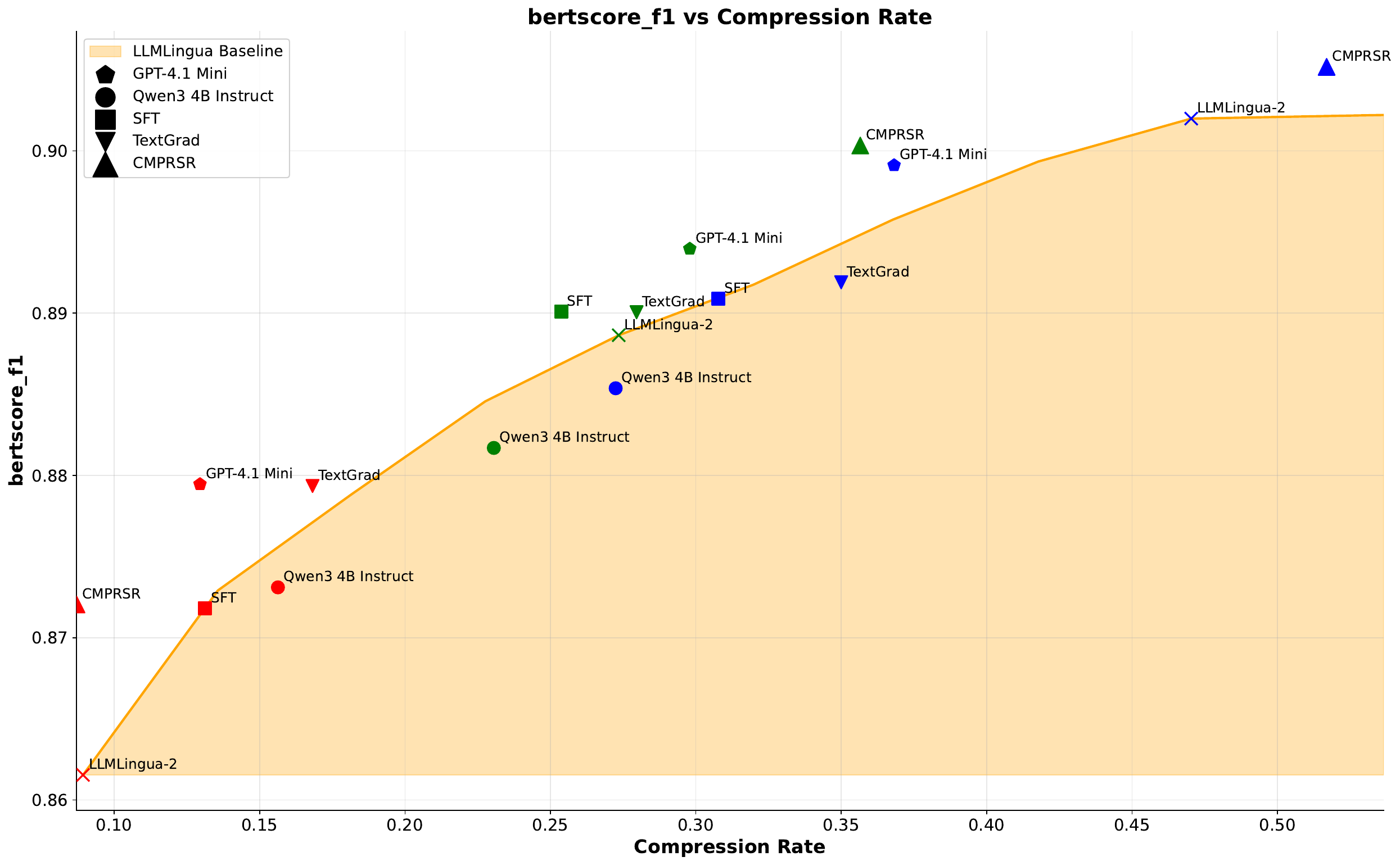}
    \caption{Comparing Cmprsr with different baselines.}
    \label{fig:vanilla_performance}
\end{figure}

Here, we demonstrate how our proposed \textbf{Compressor} surpasses the extractive baseline \textbf{LLMLingua-2}.  Figure~\ref{fig:vanilla_performance} illustrates the baseline performance curve of \textbf{LLMLingua-2} compared to several abstractive models.  When evaluated on the MeetingBank \textbf{summarization pipeline} using the \textit{BERTScore-F1} metric (on the validation split), we observe that \textit{Qwen3-4B} performs relatively poorly in terms of both semantic preservation and compression adherence.  While \textit{gpt-4.1-mini} achieves higher BERTScore than \textit{LLMLingua-2}, it still struggles to precisely follow the target CR, highlighting that large closed-source models are not ideal as standalone compressors for cost reduction.  Nevertheless, their generations can serve as valuable synthetic data for distilling compression ability into smaller models. In contrast, our proposed \textbf{Cmprsr} consistently outperforms the baselines, achieving better semantic fidelity and closely adhering to the desired CR across evaluation settings.

\subsection{Additional analysis: comparison vs.\ Lingua-2}
\label{sec:supp_analysis_gemma}

We evaluate information preservation using two complementary views: (i) cross-entropy on downstream summarization as a proxy for semantic fidelity, and (ii) surface-form retention measured via $n$-gram overlap between the compressed context and the original. Cross-entropy captures how pruning alters the model's predictive distribution, while $n$-gram overlap quantifies how much lexical material is preserved at different granularities (unigrams, bigrams, trigrams), with higher-order $n$-grams reflecting stronger phrase-level continuity.

\begin{figure*}[t!]
    \centering
    \includegraphics[width=0.9\linewidth]{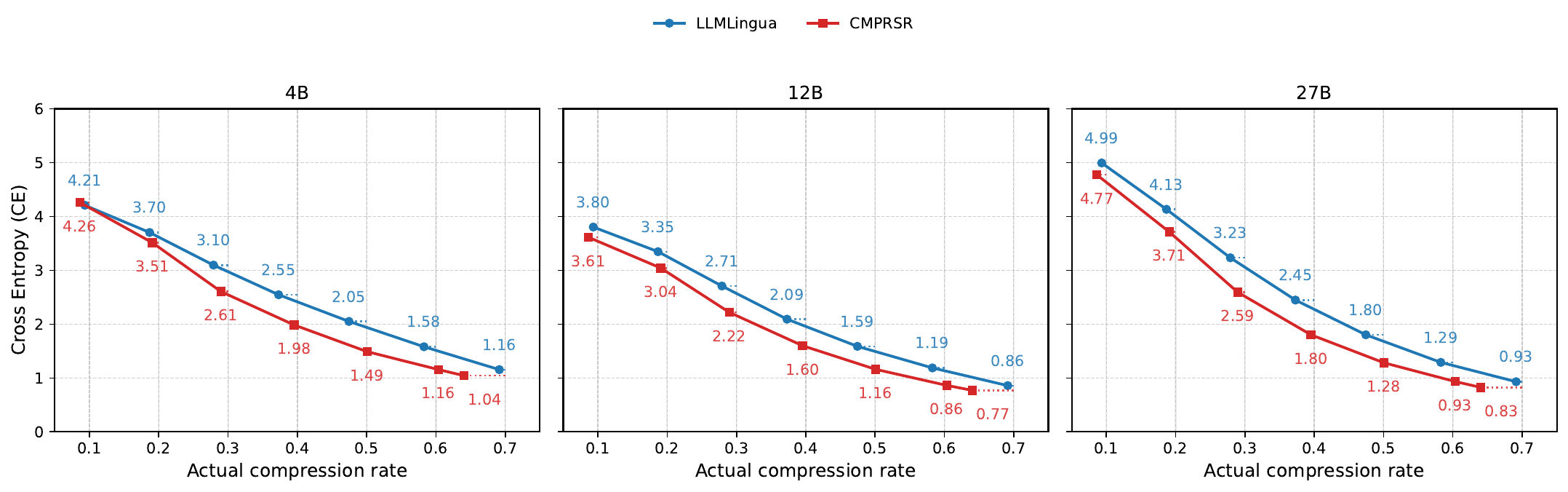}
    \caption{Cross-entropy analysis of Gemma-3 (4B, 12B, 27B) performance on MeetingBank summarization under \textsc{llmlingua-2} and \textsc{Cmprsr} compression. }
    \label{fig:ce_gemma}
\end{figure*}

In Figure~\ref{fig:ce_gemma}, we compare \textsc{LLMLingua-2} and \textsc{Cmprsr} across three model scales to identify scaling trends in robustness to context pruning. Lower cross-entropy indicates superior information retention. Results are averaged over 1{,}000 evaluation samples.

Table~\ref{tab:ngram_overlap} reports overlap statistics as a function of the target compression rate (CR). As expected, overlap increases with less aggressive compression (higher target rate) and decreases with larger $n$, since preserving longer contiguous spans is harder than preserving individual tokens. These statistics help interpret whether a method retains information primarily by keeping key words (higher 1-gram overlap) versus maintaining longer segments of the original phrasing (higher 2/3-gram overlap).

\begin{table*}[t!]
\centering
\caption{N-gram overlap statistics by target CR.}
\label{tab:ngram_overlap}
\resizebox{0.7\textwidth}{!}{%
\begin{tabular}{c cc cc cc}
\toprule
\multirow{2}{*}{\textbf{Target Rate}} & 
\multicolumn{2}{c}{\textbf{1-gram}} & 
\multicolumn{2}{c}{\textbf{2-gram}} &
\multicolumn{2}{c}{\textbf{3-gram}} \\
\cmidrule(lr){2-3} \cmidrule(lr){4-5} \cmidrule(lr){6-7}
 & \textbf{Lingua-2} & \textbf{Cmprsr} 
 & \textbf{Lingua-2} & \textbf{Cmprsr} 
 & \textbf{Lingua-2} & \textbf{Cmprsr} \\
\midrule
0.1 & 0.065 & 0.065 & 0.017 & 0.024 & 0.003 & 0.008 \\
0.2 & 0.145 & 0.141 & 0.043 & 0.055 & 0.010 & 0.021 \\
0.3 & 0.227 & 0.229 & 0.080 & 0.101 & 0.026 & 0.044 \\
0.4 & 0.318 & 0.327 & 0.137 & 0.166 & 0.057 & 0.085 \\
0.5 & 0.420 & 0.431 & 0.219 & 0.246 & 0.110 & 0.145 \\
0.6 & 0.531 & 0.535 & 0.327 & 0.343 & 0.196 & 0.227 \\
0.7 & 0.644 & 0.572 & 0.454 & 0.380 & 0.315 & 0.260 \\
\bottomrule
\end{tabular}}
\end{table*}

\section{Experimental Results} \label{A:full_results}

\subsection{Performance on QA-Specific Benchmarks}
The first set of evaluations focuses on standard Question Answering (QA) benchmarks, specifically \textbf{LongBench (LB)} and the \textbf{Ruler (R)} suite. The \textbf{Cmprsr} model demonstrates remarkable resilience under high-pressure constraints. At a 0.1 Compression Ratio (CR), Cmprsr maintains a 52.00\% precision on LB\_TriviaQA, closely trailing the uncompressed baseline (54.00\%) and outperforming GPT-4.1-mini. In the Ruler dataset, Cmprsr consistently outperform Lingua-2 at low CRs (e.g., 44.74\% vs. 41.13\% on R\_HotpotQA). These results suggest that our model effectively preserves the information required for QA retrieval even when 90\% of the original context is removed.

\subsection{Multi-Document Reasoning}
Evaluating across multiple documents in \textbf{LongBench V2 (Multi-Doc)} highlights the model's ability to synthesize information from disjoint sources. Cmprsr exhibits significant dominance in the \textbf{Multi-news} (34.78\% at 0.1 CR) and \textbf{Financial} (40.00\% at 0.1 CR) sectors. Notably, in these categories, the compressed Cmprsr model actually exceeds the precision of the "No Compression" baseline, which suggests that the compression process may act as a de-noising filter, helping the model focus on salient cross-document patterns. As the CR increases to 0.5, the model stabilizes near baseline levels for Govt. (39.13\%) and Legal (28.57\%) tasks.
\subsection{Single-Document Deep Understanding}
On \textbf{LongBench V2 (Single-Doc)} tasks, which require deep semantic tracing within a single narrative, Cmprsr proves highly adaptable. In the \textbf{Literary} and \textbf{Detective} categories, the model maintains high precision across all CR levels, reaching a peak of 36.67\% in Literary tasks at 0.3 CR. For the \textbf{Event Ordering} task, we observe a clear scaling effect: performance improves from 20.00\% to 40.00\% as the CR relaxes from 0.1 to 0.5. This indicates that while factual recall is preserved at high compression, complex chronological ordering benefits significantly from a larger token budget.

\subsection{Summarization and Document Reporting}
Finally, we assess the model’s ability to generate coherent, information-dense summaries using the $BERT_{F1}$ metric. In the \textbf{Gov. Report}, \textbf{QMSum}, and \textbf{Multi-News} benchmarks, Cmprsr achieves its highest semantic alignment at 0.5 CR, specifically reaching a $BERT_{F1}$ of 0.9089 in Multi-News. While $BERT_{F1}$ scores are inherently sensitive to compression (as shorter summaries provide fewer opportunities for token alignment with the reference), Cmprsr remains competitive with proprietary models. In the \textbf{Legal} and \textbf{Govt.}, Cmprsr demonstrates its ability to distill formal prose, matching the uncompressed baseline’s performance in the Legal domain (47.37\%) at a 0.5 CR.

\begin{table*}[htbp]
\centering
\caption{Model performance across all benchmarks under varying CRs.}
\label{tab:comprehensive_results}
\renewcommand{\arraystretch}{0.8}
% --- SUBTABLE 1: LONGBENCH & RULER ---
\begin{subtable}{\linewidth}
\centering
%\caption{LongBench (LB) \& Ruler (r) Performance}
\begin{adjustbox}{max width=\textwidth}
\begin{tabular}{c p{2.8cm} | cc cc cc | cc cc}
\toprule
\multirow{2}{*}{\textbf{CR}} & \multirow{2}{*}{\textbf{Model}} & \multicolumn{2}{c}{\textbf{LB\_TriviaQA}} & \multicolumn{2}{c}{\textbf{LB\_Qasper}} & \multicolumn{2}{c}{\textbf{LB\_NarrativeQA}} & \multicolumn{2}{c}{\textbf{R\_HotpotQA}} & \multicolumn{2}{c}{\textbf{R\_SQuAD}} \\
& & Prec. & $\Delta_{CR}$ & Prec. & $\Delta_{CR}$ & Prec. & $\Delta_{CR}$ & Prec. & $\Delta_{CR}$ & Prec. & $\Delta_{CR}$ \\
\midrule
- & \textbf{No Compression} & 0.5400 & - & 0.3400 & - & 0.4200 & - & 0.7419 & - & 0.8417 & - \\
\midrule
\multirow{4}{*}{0.1} 
& DAC & 0.5200 & 0.00 & 0.1300 & 0.00 & 0.1300 & 0.00 & 0.3536 & 0.00 & 0.2912 & -0.01 \\
& Lingua-2 & \textbf{0.5400} & \textbf{0.00} & 0.1900 & -0.02 & 0.2200 & 0.07 & 0.4113 & -0.01 & 0.3774 & -0.02 \\
& GPT-4.1-mini & 0.5000 & 0.10 & 0.3000 & 0.16 & 0.2800 & 0.09 & 0.5931 & 0.09 & 0.5089 & 0.11 \\
& \cmprsr{Cmprsr} & 0.5200 & -0.01 & \cmprsr{0.2400} & \cmprsr{0.00} & \cmprsr{0.2500} & \cmprsr{-0.01} & \cmprsr{0.4474} & \cmprsr{-0.01} & \cmprsr{0.4134} & \cmprsr{0.02} \\
\midrule
\multirow{4}{*}{0.3} 
& DAC & \textbf{0.5300} & \textbf{-0.01} & 0.2200 & -0.01 & 0.2500 & -0.01 & 0.4401 & -0.01 & 0.3995 & -0.02  \\
& Lingua-2 & 0.5100 & 0.02 & 0.2700 & -0.01 & 0.3100 & 0.01 & 0.5830 & 0.02 & 0.5981 & 0.02 \\
& GPT-4.1-mini & 0.5100 & 0.16 & 0.3400 & 0.15 & 0.3600 & 0.13 & 0.6944 & 0.19 & 0.7034 & 0.14 \\
& \cmprsr{Cmprsr} & 0.5200 & 0.03 & \cmprsr{0.3000} & \cmprsr{0.03} & \cmprsr{0.3500} & \cmprsr{0.02} & \cmprsr{0.6351} & \cmprsr{0.03} & \cmprsr{0.6049} & \cmprsr{-0.01} \\
\midrule
\multirow{4}{*}{0.5} 
& DAC & 0.4900 & -0.01 & 0.2300 & -0.02 & 0.3100 & -0.01 & 0.5149 & -0.02 & 0.4780 & -0.02\\
& Lingua-2 & \textbf{0.5500} & \textbf{0.02} & \textbf{0.3000} & \textbf{0.01} & 0.3200 & 0.04 & 0.6721 & 0.01 & 0.7365 & 0.02 \\
& GPT-4.1-mini & 0.5000 & 0.06 & 0.3300 & 0.04 & \textbf{0.3800} & \textbf{0.01} & 0.7160 & 0.09 & 0.7643 & 0.07 \\
& \cmprsr{Cmprsr} & 0.5400 & 0.04 & \cmprsr{0.3000} & \cmprsr{0.01} & 0.3600 & 0.01 & \cmprsr{0.7014} & \cmprsr{0.01} & \cmprsr{0.7371} & \cmprsr{-0.06} \\
\bottomrule
\end{tabular}
\end{adjustbox}
\end{subtable}

\bigskip

% --- SUBTABLE 2: LONGBENCH V2 MULTI-DOC ---
\begin{subtable}{\linewidth}
\centering
%\caption{LongBench V2 (Multi-Document) Performance}
\begin{adjustbox}{max width=\textwidth}
\begin{tabular}{c p{2.8cm} | cc cc cc cc | cc}
\toprule
\multirow{2}{*}{\textbf{CR}} & \multirow{2}{*}{\textbf{Model}} & \multicolumn{2}{c}{\textbf{Govt.}} & \multicolumn{2}{c}{\textbf{Legal}} & \multicolumn{2}{c}{\textbf{Multi-news}} & \multicolumn{2}{c}{\textbf{Financial}} & \multicolumn{2}{c}{\textbf{LongDialogue}} \\
& & Prec. & $\Delta_{CR}$ & Prec. & $\Delta_{CR}$ & Prec. & $\Delta_{CR}$ & Prec. & $\Delta_{CR}$ & Prec. & $\Delta_{CR}$ \\
\midrule
- & \textbf{No Compression} & 0.2174 & - & 0.3571 & - & 0.1739 & - & 0.2667 & - & 0.2100 & - \\
\midrule
\multirow{4}{*}{0.1} 
& DAC & 0.3043 & 0.00 & 0.2143 & 0.00 & 0.2174 & 0.00 & 0.2666 & 0.00 & \textbf{0.1500} & \textbf{-0.01} \\
& Lingua-2 & 0.3913 & 0.06 & \textbf{0.2857} & \textbf{-0.02} & 0.1739 & 0.06 & 0.4000 & 0.04 & 0.1200 & 0.05 \\
& GPT-4.1-mini & 0.3043 & 0.10 & 0.3571 & 0.10 & 0.2174 & 0.10 & 0.3333 & 0.10 & 0.1000 & 0.07 \\
& \cmprsr{Cmprsr} & \cmprsr{0.3913} & \cmprsr{-0.01} & 0.2143 & -0.01 & \cmprsr{0.3478} & \cmprsr{-0.01} & \cmprsr{0.4000} & \cmprsr{-0.01} & 0.1300 & -0.01 \\
\midrule
\multirow{4}{*}{0.3} 
& DAC & \textbf{0.3478} & \textbf{-0.01} & \textbf{0.2143} & \textbf{0.00} & 0.2174 & -0.01 & 0.2000 & -0.01 & 0.1000 & -0.01 \\
& Lingua-2 & 0.2609 & 0.03 & 0.1429 & 0.01 & \textbf{0.3478} & \textbf{-0.04} & 0.2000 & 0.02 & 0.1100 & -0.01 \\
& GPT-4.1-mini & 0.3043 & 0.14 & 0.3571 & 0.10 & 0.3043 & 0.17 & 0.2667 & 0.20 & 0.2000 & 0.20 \\
& \cmprsr{Cmprsr} & \cmprsr{0.3478} & \cmprsr{0.01} & \cmprsr{0.2143} & \cmprsr{0.01} & 0.2609 & 0.02 & \cmprsr{0.3333} & \cmprsr{0.02} & \cmprsr{0.1300} & \cmprsr{0.02} \\
\midrule
\multirow{4}{*}{0.5} 
& DAC & 0.3045 & -0.02 & 0.3571 & -0.01 & 0.1739 & -0.01 & 0.1333 & -0.01 & 0.1600 & -0.02 \\
& Lingua-2 & \textbf{0.3913} & \textbf{-0.02} & 0.2143 & -0.03 & 0.2609 & 0.04 & \textbf{0.2667} & \textbf{-0.02} & 0.1100 & -0.03 \\
& GPT-4.1-mini & 0.2609 & 0.05 & 0.3571 & 0.04 & 0.3478 & 0.08 & 0.3333 & 0.07 & 0.1700 & 0.11 \\
& \cmprsr{Cmprsr} & \cmprsr{0.3913} & \cmprsr{0.01} & \cmprsr{0.2857} & \cmprsr{0.00} & \cmprsr{0.2609} & \cmprsr{0.02} & \cmprsr{0.2667} & \cmprsr{0.02} & \cmprsr{0.1900} & \cmprsr{0.02} \\
\bottomrule
\end{tabular}
\end{adjustbox}
\end{subtable}
\bigskip

% --- SUBTABLE 3: LONGBENCH V2 SINGLE-DOC ---
\begin{subtable}{\linewidth}
\centering
%\caption{LongBench V2 (Single-Document) Performance}
\begin{adjustbox}{max width=\textwidth}
\begin{tabular}{c p{2.8cm} | cc cc cc cc cc}
\toprule
\multirow{2}{*}{\textbf{CR}} & \multirow{2}{*}{\textbf{Model}} & \multicolumn{2}{c}{\textbf{Literary}} & \multicolumn{2}{c}{\textbf{Academic}} & \multicolumn{2}{c}{\textbf{Detective}} & \multicolumn{2}{c}{\textbf{Event Ord.}} & \multicolumn{2}{c}{\textbf{Financial}} \\
& & Prec. & $\Delta_{CR}$ & Prec. & $\Delta_{CR}$ & Prec. & $\Delta_{CR}$ & Prec. & $\Delta_{CR}$ & Prec. & $\Delta_{CR}$ \\
\midrule
- & \textbf{No Compression} & 0.2667 & - & 0.3636 & - & 0.2273 & - & 0.2500 & - & 0.4545 & - \\
\midrule
\multirow{4}{*}{0.1} 
& DAC & 0.2600 & 0.00 & 0.3667 & 0.00 & \textbf{0.3636} & \textbf{0.00} & 0.2000 & 0.00 & 0.1364 & 0.00 \\
& Lingua-2 & 0.2667 & 0.00 & \textbf{0.4318} & \textbf{-0.03} & 0.1727 & 0.01 & \textbf{0.3000} & \textbf{-0.01} & \textbf{0.3182} & \textbf{-0.02} \\
& GPT-4.1-mini & 0.3667 & 0.07 & 0.4091 & 0.00 & 0.3182 & 0.07 & 0.1500 & 0.07 & 0.3636 & 0.11 \\
& \cmprsr{Cmprsr} & \cmprsr{0.3667} & \cmprsr{-0.01} & 0.3182 & 0.00 & 0.1818 & -0.01 & 0.2000 & -0.01 & \cmprsr{0.3182} & \cmprsr{-0.01} \\
\midrule
\multirow{4}{*}{0.3} 
& DAC & 0.2667 & -0.01 & 0.3409 & -0.01 & \textbf{0.3118} & \textbf{-0.01} & \textbf{0.3000} & \textbf{-0.01} & \textbf{0.3636} & \textbf{-0.01 }\\
& Lingua-2 & 0.2333 & 0.02 & \textbf{0.4318} & \textbf{0.01} & 0.1818 & -0.01 & \textbf{0.3000} & \textbf{0.00} & 0.2727 & -0.04 \\
& GPT-4.1-mini & 0.3333 & 0.09 & 0.3636 & 0.10 & 0.2727 & 0.09 & 0.2000 & 0.10 & 0.2273 & 0.16 \\
& \cmprsr{Cmprsr} & \cmprsr{0.3667} & \cmprsr{0.03} & 0.3636 & 0.04 & 0.2273 & 0.01 & 0.2000 & 0.01 & 0.2927 & 0.01 \\
\midrule
\multirow{4}{*}{0.5} 
& DAC & \textbf{0.3667} & \textbf{-0.01} & 0.4545 & -0.01 & 0.2273 & -0.01 & 0.2500 & -0.02 & - & - \\
& Lingua-2 & 0.2333 & -0.01 & \textbf{0.4773} & \textbf{-0.03} & 0.2273 & -0.03 & \textbf{0.4000} & \textbf{0.03} & 0.3182 & 0.03 \\
& GPT-4.1-mini & 0.3333 & -0.01 & 0.3864 & 0.05 & 0.2273 & 0.10 & 0.2000 & -0.02 & 0.4091 & 0.07 \\
& \cmprsr{Cmprsr} & 0.3000 & 0.03 & 0.4318 & 0.07 & \cmprsr{0.2727} & \cmprsr{0.02} & \cmprsr{0.4000} & \cmprsr{-0.01} & \cmprsr{0.3636} & \cmprsr{0.03} \\
\bottomrule
\end{tabular}
\end{adjustbox}
\end{subtable}
\bigskip

% --- SUBTABLE 4: SUMMARIZATION ---
\begin{subtable}{\linewidth}
\centering
%\caption{Summarization and Document Report Performance}
\begin{adjustbox}{max width=\textwidth}
\begin{tabular}{c p{2.8cm} | cc cc | cc cc cc}
\toprule
\multirow{2}{*}{\textbf{CR}} & \multirow{2}{*}{\textbf{Model}} & \multicolumn{2}{c}{\textbf{Govt.}} & \multicolumn{2}{c}{\textbf{Legal}} & \multicolumn{2}{c}{\textbf{QMSum}} & \multicolumn{2}{c}{\textbf{Multi-news}} & \multicolumn{2}{c}{\textbf{Gov. Report}} \\
& & Prec. & $\Delta_{CR}$ & Prec. & $\Delta_{CR}$ & BERT$_{F1}$ & $\Delta_{CR}$ & BERT$_{F1}$ & $\Delta_{CR}$ & BERT$_{F1}$ & $\Delta_{CR}$ \\
\midrule
- & \textbf{No Compression} & 0.3333 & - & 0.4737 & - & 0.8900 & - & 0.9308 & - & 0.9023 & - \\
\midrule
\multirow{4}{*}{0.1} 
& DAC & 0.2222 & 0.00 & \textbf{0.4737} & \textbf{0.00} & 0.8503 & 0.00 & 0.8575 & 0.00 & 0.8500 & 0.00 \\
& Lingua-2 & \textbf{0.2778} & \textbf{-0.01} & 0.3684 & -0.02 & 0.8596 & 0.04 & 0.8580 & -0.02 & 0.8527 & -0.02 \\
& GPT-4.1-mini & 0.3333 & 0.13 & 0.4737 & 0.06 & 0.8663 & 0.08 & 0.8852 & 0.05 & 0.8723 & 0.06 \\
& \cmprsr{Cmprsr} & 0.2667 & -0.01 & 0.3684 & -0.01 & \cmprsr{0.8635} & \cmprsr{0.00} & \cmprsr{0.8815} & \cmprsr{0.00} & \cmprsr{0.8645} & \cmprsr{-0.01} \\
\midrule
\multirow{4}{*}{0.3} 
& DAC & 0.2778 & -0.01 & 0.4211 & -0.01 & 0.8641 & -0.01 & 0.8778 & -0.01 & 0.8628 & -0.01 \\
& Lingua-2 & 0.3333 & -0.05 & 0.4211 & 0.02 & 0.8679 & -0.01 & 0.8887 & 0.03 & 0.8748 & 0.06 \\
& GPT-4.1-mini & 0.2222 & 0.19 & 0.5789 & 0.12 & 0.8704 & 0.14 & 0.8955 & 0.05 & 0.8853 & 0.09 \\
& \cmprsr{Cmprsr} & \cmprsr{0.3667} & \cmprsr{0.02} & \cmprsr{0.5789} & \cmprsr{0.03} & \cmprsr{0.8761} & \cmprsr{-0.04} & \cmprsr{0.8995} & \cmprsr{0.01} & \cmprsr{0.8862} & \cmprsr{0.04} \\
\midrule
\multirow{4}{*}{0.5} 
& DAC & \textbf{0.3333} & \textbf{-0.01} & 0.3158 & -0.01 & 0.8737 & -0.01 & 0.8941 & -0.01 & 0.8756 & -0.01 \\
& Lingua-2 & 0.2778 & 0.02 & 0.4211 & -0.02 & 0.8780 & 0.02 & 0.9084 & 0.03 & 0.8829 & 0.03 \\
& GPT-4.1-mini & 0.2222 & 0.11 & 0.3684 & 0.05 & 0.8714 & 0.04 & 0.9004 & -0.08 & 0.8875 & 0.01 \\
& \cmprsr{Cmprsr} & \cmprsr{0.3333} & \cmprsr{0.03} & \cmprsr{0.4737} & \cmprsr{0.02} & \cmprsr{0.8814} & \cmprsr{-0.06} & \cmprsr{0.9089} & \cmprsr{0.00} & \cmprsr{0.8929} & \cmprsr{0.08} \\
\bottomrule
\end{tabular}
\end{adjustbox}
\end{subtable}
\end{table*}

\begin{table*}[t!]
\centering
\caption{Model performance across Question Answering and Summarization benchmarks. As our objective is to maximize task performance under a relaxed constraint of $\Delta_{CR} \leq \epsilon$, we focus on the models that respect the CR budget and highlight the best-performing method within the corresponding subset. We set $\epsilon$ to 0.03. Note that perfect adherence to the CR target is often unattainable, and a slightly negative $\Delta_{CR}$ is preferable. }
\label{tab:full_results}
\renewcommand{\arraystretch}{0.78}

% --- SUBTABLE 1 ---
\begin{subtable}{\linewidth}
\centering
\caption{Performance on InfiniteBench, LongBench V2, and Ruler}
\begin{adjustbox}{max width=\textwidth}
\begin{tabular}{p{0.5cm} p{2.8cm} | cc cc cc cc}
\toprule
\multirow{2}{*}{\textbf{CR}} & \multirow{2}{*}{\textbf{Model}} & \multicolumn{2}{c}{\textbf{LongBench QA}} & \multicolumn{2}{c}{\textbf{InfiniteBench}} & \multicolumn{2}{c}{\textbf{LongBench V2}} & \multicolumn{2}{c}{\textbf{Ruler}} \\
& & Prec. & $\Delta_{CR}$ & Prec. & $\Delta_{CR}$ & Prec. & $\Delta_{CR}$ & Prec. & $\Delta_{CR}$ \\
\midrule
- & No Compression & 0.4333 & - & 0.2100 & - & 0.3384 & - & 0.7918 & - \\
\midrule
\multirow{4}{*}{0.1} 
& DAC & 0.2600 & 0.00 & \textbf{0.1500} & \textbf{-0.01} & 0.2886 & 0.00 & 0.3224 & -0.01 \\
& Lingua-2 & 0.3167 & 0.02 & 0.1200 & 0.05 & \textbf{0.3051} & \textbf{-0.01} & 0.3944 & -0.02 \\
& GPT-4.1-mini & 0.3600 & 0.12 & 0.1000 & 0.07 & 0.3449 & 0.07 & 0.5510 & 0.10 \\
& \cmprsr{Cmprsr} & \cmprsr{{0.3367}} & \cmprsr{{-0.01}} & 0.1300 & -0.01 & 0.2886 & -0.01 & \cmprsr{{0.4304}} & \cmprsr{{0.01}} \\
\midrule
\multirow{4}{*}{0.3} 
& DAC & 0.3333 & -0.01 & 0.1000 & -0.01 & 0.3260 & -0.01 & 0.4198 & -0.02 \\
& Lingua-2 & 0.3633 & 0.01 & 0.1100 & -0.01 & 0.3106 & -0.01 & 0.5905 & 0.02 \\
& GPT-4.1-mini & 0.4033 & 0.15 & 0.2000 & 0.20 & 0.3140 & 0.12 & 0.6989 & 0.17 \\
& \cmprsr{Cmprsr} & \cmprsr{{0.3900}} & \cmprsr{{0.03}} & \cmprsr{{0.1300}} & \cmprsr{{0.02}} & \cmprsr{{0.3423}} & \cmprsr{{0.02}} & \cmprsr{{0.6200}} & \cmprsr{{0.02}} \\
\midrule
\multirow{4}{*}{0.5} 
& DAC & 0.3433 & -0.01 & 0.1600 & -0.02 & 0.3367 & -0.01 & 0.4965 & -0.02 \\
& Lingua-2 & 0.3900 & 0.02 & 0.1100 & -0.03 & 0.3364 & 0.00 & 0.7043 & 0.01 \\
& GPT-4.1-mini & 0.4033 & 0.04 & 0.1700 & 0.11 & 0.3067 & 0.05 & 0.7402 & 0.08 \\
& \cmprsr{Cmprsr} & \cmprsr{{0.4000}} & \cmprsr{{0.03}} & \cmprsr{{0.1900}} & \cmprsr{{0.02}} & \cmprsr{{0.3679}} & \cmprsr{{0.03}} & \cmprsr{{0.7193}} & \cmprsr{{0.00}} \\
\bottomrule
\end{tabular}
\end{adjustbox}
\end{subtable}

\bigskip

% --- SUBTABLE 2 ---
\begin{subtable}{\linewidth}
\centering
\caption{Performance on GSM8K, MeetingBank, and LongBench}
\begin{adjustbox}{max width=\textwidth}
\begin{tabular}{p{0.5cm} p{2.8cm} | cc cc | cc cc}
\toprule
\multirow{2}{*}{\textbf{CR}} & \multirow{2}{*}{\textbf{Model}} & \multicolumn{2}{c}{\textbf{GSM8K}} & \multicolumn{2}{c}{\textbf{MeetingBank QA}} & \multicolumn{2}{c}{\textbf{MeetingBank}} & \multicolumn{2}{c}{\textbf{LongBench}} \\
& & Prec. & $\Delta_{CR}$ & Prec. & $\Delta_{CR}$ & BERT$_{F1}$ & $\Delta_{CR}$ & BERT$_{F1}$ & $\Delta_{CR}$ \\
\midrule
- & No Compression & 0.9300 & - & 0.8410 & - & 0.9212 & - & 0.9077 & - \\
\midrule
\multirow{4}{*}{0.1} 
& DAC & 0.0000 & -0.02 & 0.2410 & -0.01 & 0.8562 & -0.01 & 0.8500 & 0.00 \\
& Lingua-2 & 0.0033 & 0.04 & \textbf{0.4475} & \textbf{0.01} & 0.8641 & -0.02 & 0.8568 & 0.00 \\
& GPT-4.1-mini & 0.4933 & 0.12 & 0.5030 & 0.03 & 0.8824 & 0.03 & 0.8746 & 0.07 \\
& \cmprsr{Cmprsr} & \cmprsr{0.1300} & \cmprsr{0.03} & 0.4430 & 0.00 & \cmprsr{0.8775} & \cmprsr{0.00} & \cmprsr{0.8698} & \cmprsr{0.00} \\
\midrule
\multirow{4}{*}{0.3} 
& DAC & 0.0034 & -0.04 & 0.3815 & 0.00 & 0.8738 & 0.00 & 0.8682 & -0.01 \\
& Lingua-2 & 0.1200 & 0.03 & 0.6810 & -0.01 & 0.8894 & -0.01 & 0.8771 & 0.03 \\
& GPT-4.1-mini & 0.8067 & 0.09 & 0.6970 & -0.01 & 0.8953 & -0.01 & 0.8837 & 0.09 \\
& \cmprsr{Cmprsr} & \cmprsr{0.4667} & \cmprsr{0.03} & \cmprsr{0.6880} & \cmprsr{0.02} & \cmprsr{{0.9033}} & \cmprsr{0.02} & \cmprsr{0.8872} & \cmprsr{0.00} \\
\midrule
\multirow{4}{*}{0.5} 
& DAC & 0.0300 & -0.06 & 0.4840 & -0.01 & 0.8873 & -0.01 & 0.8811 & -0.01 \\
& Lingua-2 & 0.5133 & 0.02 & \textbf{0.7855} & \textbf{0.02} & 0.9024 & 0.02 & 0.8898 & 0.03 \\
& GPT-4.1-mini & 0.9100 & 0.07 & 0.7460 & -0.14 & 0.9003 & -0.14 & 0.8864 & -0.01 \\
& \cmprsr{Cmprsr} & \cmprsr{0.7200} & \cmprsr{-0.05} & 0.7670 & 0.02 & \cmprsr{0.9106} & \cmprsr{0.02} & \cmprsr{0.8944} & \cmprsr{0.01} \\
\bottomrule
\end{tabular}
\end{adjustbox}
\end{subtable}
\end{table*}

\section{Compression Examples}

\subsection{Meeting Transcript}

\textbf{Original Text: } Good morning everyone. Let's start today's quarterly review meeting. First on the agenda is the sales performance for Q3. Sarah, can you walk us through the numbers? Thank you, John. Yes, so we saw a 15\% increase in overall sales compared to Q2. Our enterprise division performed exceptionally well with a 23\% growth, driven primarily by our new cloud solutions package. The SMB segment grew by 8\%, which is slightly below our target of 10\%. We believe this is due to increased competition in that space. Our customer acquisition cost decreased by 12\%, which is excellent news for our margins. However, our churn rate increased slightly from 2.1\% to 2.4\%, something we need to address urgently. Thank you Sarah. Those are solid numbers overall. Mark, what's your take on the churn rate increase? Well, from our customer success team's perspective, we've identified three main factors. First, some customers found our recent UI updates confusing and felt they weren't adequately trained. Second, we had some service disruptions in August that affected about 200 enterprise clients. Third, our competitor launched an aggressive pricing campaign targeting our mid-tier customers. We've already started implementing solutions. We're rolling out comprehensive training webinars, we've improved our system reliability with additional redundancy, and we're preparing a value-add package to counter the competitive pressure. That sounds good Mark. What's the timeline on these initiatives? The training program launches next week, the infrastructure improvements went live yesterday, and the value-add package will be ready by end of month. Excellent. Next item is product development. Lisa, where are we with the new AI features? Thanks John. We're making great progress on the AI-powered analytics module. The beta version has been testing internally for three weeks now with positive feedback. We plan to release it to select customers in November for pilot testing. The natural language query feature is particularly exciting - users can now ask questions in plain English and get instant insights from their data. We've also enhanced our API based on customer requests. The team has been working overtime to meet the deadlines, and I'm confident we'll deliver on schedule. That's fantastic Lisa. Now let's discuss marketing strategy for Q4.

\textbf{Cmprsr (Target 0.3 / Actual 0.3556)}: Q3 sales 15\% increase overall. Enterprise division exceptional 23\% growth driven new cloud solutions package. SMB segment grew 8\% below target 10\% due competition. Customer acquisition cost decreased 12\% good margins. Churn rate increased from 2.1\% to 2.4\% urgent. Mark customer success team identifies three factors: customers found recent UI updates confusing untrained. Service disruptions August affecting 200 enterprise clients. Competitor aggressive pricing campaign mid-tier. We rolling out comprehensive training webinars next week, infrastructure improvements live yesterday, value-add package ready end month. Product development new AI features: AI-powered analytics module beta testing internally positive feedback plan release select customers November pilot. Natural language query feature exciting users ask plain English get insights. Enhanced API capabilities.

\textbf{LLM-Lingua (Target 0.3 / Actual 0.3039):} morning quarterly review meeting sales performance Q3 Sarah? 15\% increase sales Q2. enterprise division 23\% growth cloud solutions SMB segment grew 8\% below target 10\% increased competition customer cost decreased 12\% churn rate increased 2.1\% to 2.4\% churn rate increase? three factors UI updates confusing service disruptions August 200 enterprise clients competitor pricing campaign mid-tier implementing solutions training webinars improved system reliability preparing value-add package? training infrastructure improvements value-add package end month product development new AI features? progress AI-powered module beta version positive feedback release November natural language query feature insights enhanced API capabilities team working overtime deadlines deliver schedule marketing Q4.

\textbf{DAC (Target 0.3 / Actual 0.2950):} Good. start quarterly the the sales for3 Sarah the numbers?, John Yes so saw5 increase sales compared2 Our enterprise division performed growth our package The SMB grew slightly target\%. believe due competition space customer decreased \%, excellent margins However churn\%, something address urgently Thank Sarah Those solid Mark churn Well from success's identified three factors some found recent UI confusing felt weren adequately Second service disruptions August about enterprise our competitor an targeting've implementing rolling comprehensive web system additional're value competitive That sounds Mark timeline initiatives The program launches week infrastructure went yesterday ready Excellent Next item product Lisa where AI Thanks We making-powered beta testing plan select November natural query particularly ask instant've API requests overtime meet deadlines confident fantastic discuss.

\textbf{Cmprsr (Target 0.5 / Actual 0.5539)} Good morning quarterly review meeting. First sales performance Q3. Sarah please. We saw 15\% increase overall sales compared Q2. Enterprise division performed exceptionally well 23\% growth driven by new cloud solutions package. SMB segment grew 8\% below target 10\% due increased competition. Customer acquisition cost decreased by 12\% good for margins. However churn rate increased slightly from 2.1\% to 2.4\% need urgent address. Mark your take. Customer success team identifies three factors: First some customers found recent UI updates confusing not adequately trained. Second service disruptions in August affected about 200 enterprise clients. Third competitor launched aggressive pricing campaign targeting mid-tier customers. We already implementing solutions: rolling out comprehensive training webinars, improved system reliability with additional redundancy, preparing value-add package counter competition. Timeline: training program launches next week, infrastructure improvements went live yesterday, value-add package ready by end month. Next product development. Lisa new AI features. Great progress AI-powered analytics module beta testing internally three weeks positive feedback will release to select customers November pilot. Natural language query feature exciting users can ask plain English get instant insights. Enhanced API capabilities based customer requests. Team working overtime confident on schedule. Now discuss marketing strategy Q4.

\textbf{LLM-Lingua (Target 0.5 / Actual 0.4698):} morning start today quarterly review meeting agenda sales performance for Q3 Sarah walk through numbers? John 15\% increase sales compared to Q2. enterprise division 23\% growth new cloud solutions package SMB segment grew 8\% below target 10\% due to increased competition customer cost decreased 12\% excellent news for margins churn rate increased from 2.1\% to 2.4\% need address urgently solid numbers Mark take churn rate increase? customer success team identified three factors customers found UI updates confusing trained service disruptions August affected 200 enterprise clients competitor launched aggressive pricing campaign mid-tier customers started implementing solutions rolling training webinars improved system reliability redundancy preparing value-add package counter competitive pressure timeline initiatives? training program launches next week infrastructure improvements live value-add package ready end of month Next product development Lisa new AI features? progress on AI-powered analytics module beta version testing three weeks positive feedback plan release customers in November pilot testing natural language query feature exciting ask questions plain English instant insights data enhanced API capabilities customer requests team working overtime meet deadlines confident deliver on schedulemarketing strategy Q4

\textbf{DAC (Target 0.5 / Actual 0.4940):} Good morning everyone. Let start today quarterly review meeting. First on the is the sales performance for3. Sarah, walk through the numbers? Thank, John Yes so we saw increase overall sales compared2. Our enterprise division performed exceptionally with2 growth, driven our cloud solutions package. The SMB grew8 which slightly target of\%. We believe due competition that space. customer decreased \%, excellent margins However churn increased from  something address urgently Thank Sarah Those solid overall. Mark what take the churn increase Well from customer success identified three factors., some found our recent UI confusing felt weren adequately Second we service disruptions August that about enterprise. our competitor launched an campaign targeting mid. We've implementing solutions. rolling comprehensive web, we improved system with additional're preparing value-add package competitive That sounds Mark What timeline initiatives The program launches week infrastructure went yesterday package ready month Excellent Next item product Lisa where with the AI features Thanks We making the-powered analytics module. beta has testing internally with. plan release select November for The natural query particularly - users ask English instant insights.'ve enhanced API capabilities based customer requests The team overtime meet deadlines I confident schedule That fantastic Now discuss for.

\newpage
\subsection{Customer Service Call}

\textbf{Original Text:}
Thank you for calling TechSupport Solutions, my name is Jennifer, how may I assist you today? Hi Jennifer, I'm having a really frustrating problem with my internet connection. It keeps dropping every few minutes and I work from home so this is causing major issues. I've already tried restarting my router multiple times but nothing seems to work. I'm sorry to hear you're experiencing connectivity issues. I understand how disruptive this can be, especially when you're working from home. Let me help you resolve this. Can I start by getting your account number or the phone number associated with your account? Sure, it's 555-0123. Thank you. I'm pulling up your account now. I can see you're on our Premium Fiber plan with download speeds up to 1 gigabit. Can you tell me when this issue started? It started about three days ago, on Monday morning. Before that everything was working perfectly fine. I see. And when the connection drops, does it come back on its own or do you need to restart the router? It usually comes back after a minute or two, but sometimes I have to unplug the router and plug it back in. Okay, that's helpful information. I'm running a diagnostic test on your line right now. Can you tell me what color lights are showing on your router at the moment? The power light is solid green, the internet light is blinking amber, and the WiFi light is solid green. Alright, the blinking amber on the internet light indicates an unstable connection. I'm seeing some unusual signal fluctuations on our end. Let me check if there are any service outages in your area. What's your zip code? It's 90210. Thank you. I'm not showing any reported outages in your area, but I am seeing that your signal strength has been degrading over the past week. This could be due to a few factors - possibly a damaged cable, interference, or an issue with the router itself. Your router is about four years old, which is reaching the end of its optimal lifespan. Here's what I'd like to do. I'm going to schedule a technician to come to your location to inspect the line and test the equipment. There's no charge for this service visit. We have availability tomorrow between 1 PM and 5 PM, or Friday morning between 8 AM and noon. Which works better for you? Tomorrow afternoon would be perfect. I'll be working from home anyway.

\textbf{Cmprsr (Target 0.3 / Actual 0.3438):}
Thank you Jennifer TechSupport Solutions. I frustrated internet connection keeps dropping every few minutes affecting work from home. Tried restarting router multiple times. Your account number 555-0123. You on Premium Fiber plan up to 1 gigabit. Started three days ago Monday morning before fine. When drops does come back on own or need restart? Usually comes back after minute or sometimes unplug router. Router lights: power solid green, internet blinking amber, WiFi solid green. Blinking amber indicates unstable connection. No reported outages in your zip 90210 but signal strength degrading past week. Could due damaged cable interference or router four years old nearing end lifespan. I will schedule technician visit inspect line and test equipment no charge. Availability tomorrow 1 PM to 5 PM or Friday morning 8 AM noon. Tomorrow afternoon perfect.

\textbf{LLM-Lingua (Target 0.3 / Actual 0.2888):}
calling TechSupport Solutions Jennifer assist? frustrating problem internet connection dropping work from home major issues tried restarting router connectivity issues disruptive working from home help resolve account number? 555-0123 Premium Fiber plan download speeds 1 gigabit issue started? three days ago Monday connection drops back restart router? unplug plug running diagnostic test color lights router? power light solid green internet light blinking amber WiFi light solid green blinking amber unstable connection unusual signal fluctuations service outages zip code? 90210 reported outages signal strength degrading due to damaged cable interference issue router router four years old end optimal lifespan schedule technician inspect test equipment no charge service visitavailability tomorrow 1 PM 5 PM Friday 8 AM? home

\textbf{DAC(Target 0.3 / Actual 0.2950):} Thank calling TechSupport Solutions, my is Jennifer, assist? Hi having really frustrating problem internet connection. dropping minutes I work home so causing.'ve restarting router multiple but work sorry hear experiencing connectivity. understand disruptive you working Let resolve. start getting account number or associated? Sure it -. Thank pulling now our Premium Fiber plan download1 gig Can when issue started three ago on Before everything And drops does own restart the? minute un Okay helpful running diagnostic line tell color lights your at the? solid internet blinking amber WiFi Alright the indicates an seeing unusual our Let there serviceages area What zip9 Thank not reportedages am signal de week This due few possibly damaged interference Your four reaching Here'd'm schedule location inspect There charge We availability between PM5 Which anyway. 
        
\textbf{Cmprsr (Target 0.5 / Actual 0.4971):}
Thank you calling TechSupport Solutions I am Jennifer how assist you. I having frustrating problem my internet connection keeps dropping every few minutes and I work from home so major issues. I already tried restarting router multiple times no work. Can I your account number or phone number? Sure 555-0123. I see you on Premium Fiber plan with download speeds up to one gigabit. When this started? Three days ago Monday morning before fine. When drops does it come back on own or need restart router? Usually comes back after minute or two but sometimes must unplug replug. I running diagnostic test your line. What lights on router currently? Power light solid green, internet light blinking amber, WiFi light solid green. Blinking amber indicates unstable connection. Seeing unusual signal fluctuations. Any service outages in your area? Your zip code 90210. No reported outages but your signal strength has degraded past week. Could due damaged cable interference or router issue. Your router about four years old nearing end optimal lifespan. I will schedule technician come inspect line and test equipment. No charge. Availability tomorrow between 1 PM to 5 PM or Friday morning 8 AM to noon. Tomorrow afternoon perfect I working home anyway.

\textbf{LLM-Lingua (Target 0.5 / Actual 0.4695):}
you for calling TechSupport Solutions name Jennifer assist you today? Jennifer frustrating problem with internet connection keeps dropping every few minutes work from home causing major issues tried restarting router multiple times nothing work sorry experiencing connectivity issues understand disruptive especially working from home help resolve. start by your account number phone number? it's 555-0123. pulling up account on Premium Fiber plan with download speeds up to 1 gigabit. issue started? started three days ago Monday morning Before everything was working fine connection drops come back on or need restart router? usually comes back after minute or two sometimes unplug router plug back in information running diagnostic test on line color lights on router? power light solid green internet light blinking amber WiFi light solid green blinking amber on internet light indicates unstable connection unusual signal fluctuations check if service outages in area. zip code? 90210. not showing reported outages signal strength degrading past week due to damaged cable interference or issue with router router four years old reaching end of optimal lifespan schedule technician to inspect line test equipment no charge for service visit.availability tomorrow 1 PM 5 PM Friday 8 AM noon works? Tomorrow afternoon perfect working from home

\textbf{DAC (Target 0.5 / Actual 0.4930):} Thank you calling TechSupport Solutions, my is Jennifer, how may assist? Hi Jennifer, I having a really frustrating problem my internet connection. It keeps dropping minutes and I work home so this causing. I've restarting router multiple but to work. I sorry hear experiencing connectivity. I understand disruptive, you working home Let resolve. Can I start your account number or the phone associated your? Sure it -0. Thank. pulling now can our Premium Fiber plan with download1 gig. Can tell when this issue started? It three ago, on Monday Before everything working perfectly fine I see And when the drops, does come back own or restart the? It usually after minute two I to un the. Okay that helpful information running diagnostic test line right. Can tell color lights showing your at the? power solid internet blinking amber WiFi. Alright the on indicates an unstable. seeing some unusual signal fluctuations our end Let check there serviceages area What zip? It9. Thank not showing reportedages area, am seeing signal has de the week This due few - possibly damaged interference an router Your about four, reaching optimal Here what'd'm schedule technician come location line There no charge service visit We availability between1 PM5, morning8 noon Which? would perfect be anyway
\newpage

\subsection{Legal Document Chunk}

\textbf{Original Text:}
Section 42 - Privacy Rights and Data Protection Obligations. This section establishes comprehensive requirements for entities that collect, process, store, or transmit personal information of California residents. For purposes of this statute, 'personal information' means information that identifies, relates to, describes, is reasonably capable of being associated with, or could reasonably be linked, directly or indirectly, with a particular consumer or household. Personal information includes, but is not limited to: identifiers such as real name, alias, postal address, unique personal identifier, online identifier, Internet Protocol address, email address, account name, social security number, driver's license number, passport number, or other similar identifiers; commercial information including records of personal property, products or services purchased, obtained, or considered, or other purchasing or consuming histories or tendencies; biometric information including imagery of the iris, retina, fingerprint, face, hand, palm, vein patterns, and voice recordings, keystroke patterns or rhythms, gait patterns or rhythms, and sleep, health, or exercise data that contain identifying information; internet or other electronic network activity information including, but not limited to, browsing history, search history, and information regarding a consumer's interaction with an Internet Web site, application, or advertisement; geolocation data; audio, electronic, visual, thermal, olfactory, or similar information; professional or employment-related information; education information that is not publicly available personally identifiable information as defined in the Family Educational Rights and Privacy Act. A business that controls the collection of a consumer's personal information shall implement reasonable security procedures and practices appropriate to the nature of the personal information to protect the personal information from unauthorized or illegal access, destruction, use, modification, or disclosure. In the event of a breach of the security of the system, the business shall notify affected California residents without unreasonable delay and in no case later than forty-five days from the discovery of the breach. The notification shall include the nature of the breach, the types of information that were compromised, the actions taken by the business to remediate the breach, contact information for the business, and information about what steps consumers can take to protect themselves from potential harm.

\textbf{Cmprsr (Target 0.3 / Actual 0.3034):}
Section 42 Privacy Rights California residents. Personal information means identifies or reasonably linked particular consumer or household including identifiers like name address SS number etc; commercial information purchasing history; biometric information like fingerprint face voice keystroke gait sleep health data; internet electronic network activity browsing search interaction; geolocation data; audio visual thermal olfactory; professional employment education information not publicly available FEPA. Business controlling collection must implement reasonable security procedures protect personal information from unauthorized access destruction use modification disclosure. In breach must notify affected California residents without unreasonable delay no later than forty-five days including nature of breach types compromised information actions remediated contact business and steps consumers can protect themselves.

\textbf{LLM-Lingua (Target 0.3 / Actual 0.2644):}
Section 42 Privacy Rights Data Protection Obligations establishes requirements entities personal information California residents information consumer household information includes name alias postal identifier online identifier Internet Protocol email account name social security driver's license passport commercial information property products services purchasing histories biometric information keystroke gait sleep health exercise data internet activity information browsing geolocation data audio electronic visual thermal olfactory information professional employment-related information education information identifiable Family Educational Rights Privacy Act business information security procedures protect unauthorized access destruction use modification disclosure breach residents forty-five days from discoverynotification nature breach types information compromised actions contact information steps consumers protect harm

\textbf{DAC (Target 0.3 / Actual 0.2940):} Section  - Privacy Rights Data Oblations. This establishes comprehensive requirements entities that collect personal information California. For statute ' information means identifies reasonably associated linked consumer household. Personal includes limited: identifiers real, alias postal unique online Internet account social, driver identifiers; commercial information including records personal property products obtained considered other purchasing consuming histories tendenciesometric including imagery of the iris fingerprint hand palm, keyst patterns rhythms g patterns sleep exercise that contain; other network limited browsing regarding interaction an Web advertisement ge data electronic thermal ol similar professional employment that not personally Family controls collection a shall implement security appropriate from illegal destruction use In system notify without no than discovery notification include types, about.

\textbf{Cmprsr (Target 0.5 / Actual 0.4690):}
Section 42 - Privacy Rights and Data Protection Obligations. This establishes comprehensive requirements for entities that collect process store or transmit personal information of California residents. Personal information means information that identifies or reasonably capable associated with particular consumer or household. Includes identifiers like real name postal address unique personal identifier online identifier IP email account social security driver license passport; commercial information purchasing histories; biometric information like iris retina fingerprint face hand palm vein patterns voice keystroke gait sleep health exercise data; internet electronic network activity browsing search interaction website application advertisement; geolocation data; audio visual thermal olfactory information; professional employment related information education information not publicly available as per Family Educational Rights and Privacy Act. Business controlling collection shall implement reasonable security procedures appropriate nature of information to protect from unauthorized illegal access destruction use modification or disclosure. In breach of security must notify affected California residents without unreasonable delay no later than forty-five days. Notification shall include nature of breach types compromised information actions remediated breach business contact information and steps consumers can take protect themselves.

\textbf{LLM-Lingua (Target 0.5 / Actual 0.4345):}
Section 42 - Privacy Rights Data Protection Obligations section establishes requirements for entities collect process store transmit personal information California residents 'personal information means identifies relates consumer household Personal information includes real name alias postal address unique personal identifier online identifier Internet Protocol address email address account name social security number driver's license number passport number identifiers commercial information records personal property products services purchased purchasing consuming histories tendencies biometric information imagery iris retina fingerprint face hand palm vein patterns voice recordings keystroke gait sleep health exercise data identifying internet electronic network activity information browsing history search history geolocation data audio electronic visual thermal olfactory information professional employment-related information education information not publicly available personally identifiable information Family Educational Rights Privacy Act business controls collection personal information implement security procedures practices protect from unauthorized illegal access destruction use modification disclosure breach security business notify California residents without delay forty-five days from discovery breachnotification include nature breach types information compromised actions business to remediate breach contact information information steps consumers to protect from potential harm.

\textbf{DAC (Target 0.5 / Actual 0.4689):} Section 42 - Privacy Rights and Data Protection Obligations. This establishes comprehensive requirements for entities that collect process store or transmit personal information of California residents. Personal information means information that identifies or reasonably capable associated with particular consumer or household. Includes identifiers like real name postal address unique personal identifier online identifier IP email account social security driver license passport; commercial information purchasing histories; biometric information like iris retina fingerprint face hand palm vein patterns voice keystroke gait sleep health exercise data; internet electronic network activity browsing search interaction website application advertisement; geolocation data; audio visual thermal olfactory information; professional employment related information education information not publicly available as per Family Educational Rights and Privacy Act. Business controlling collection shall implement reasonable security procedures appropriate nature of information to protect from unauthorized illegal access destruction use modification or disclosure. In breach of security must notify affected California residents without unreasonable delay no later than forty-five days. Notification shall include nature of breach types compromised information actions remediated breach business contact information and steps consumers can take protect themselves.

% Rewrite:
\newpage
\section{TextGrad Implementation Details}
\label{textgrad-appendix}

We utilize \textsc{TextGrad} to optimize the compressor's \emph{system prompt} using question answering (QA) on the MeetingBank dataset \cite{hu-etal-2023-meetingbank} as the source of the learning signal as follows:
\begin{enumerate}
    \item Run the end-to-end pipeline on a batch of transcripts and score outputs against the ground truth;
    \item Ask an optimizer-LLM to analyze errors in the output and generate improvement suggestions;
    \item Propagate this feedback to the upstream nodes and propose a revised prompt;
    \item Accept the updated prompt if it outperforms the initial prompt on a hold-out set;
    \item Re-evaluate the pipeline with the updated prompt and repeat until reaching the budget constraints.
\end{enumerate}

\begin{figure}[h!]
  \centering
  \includegraphics[width=0.9\linewidth]{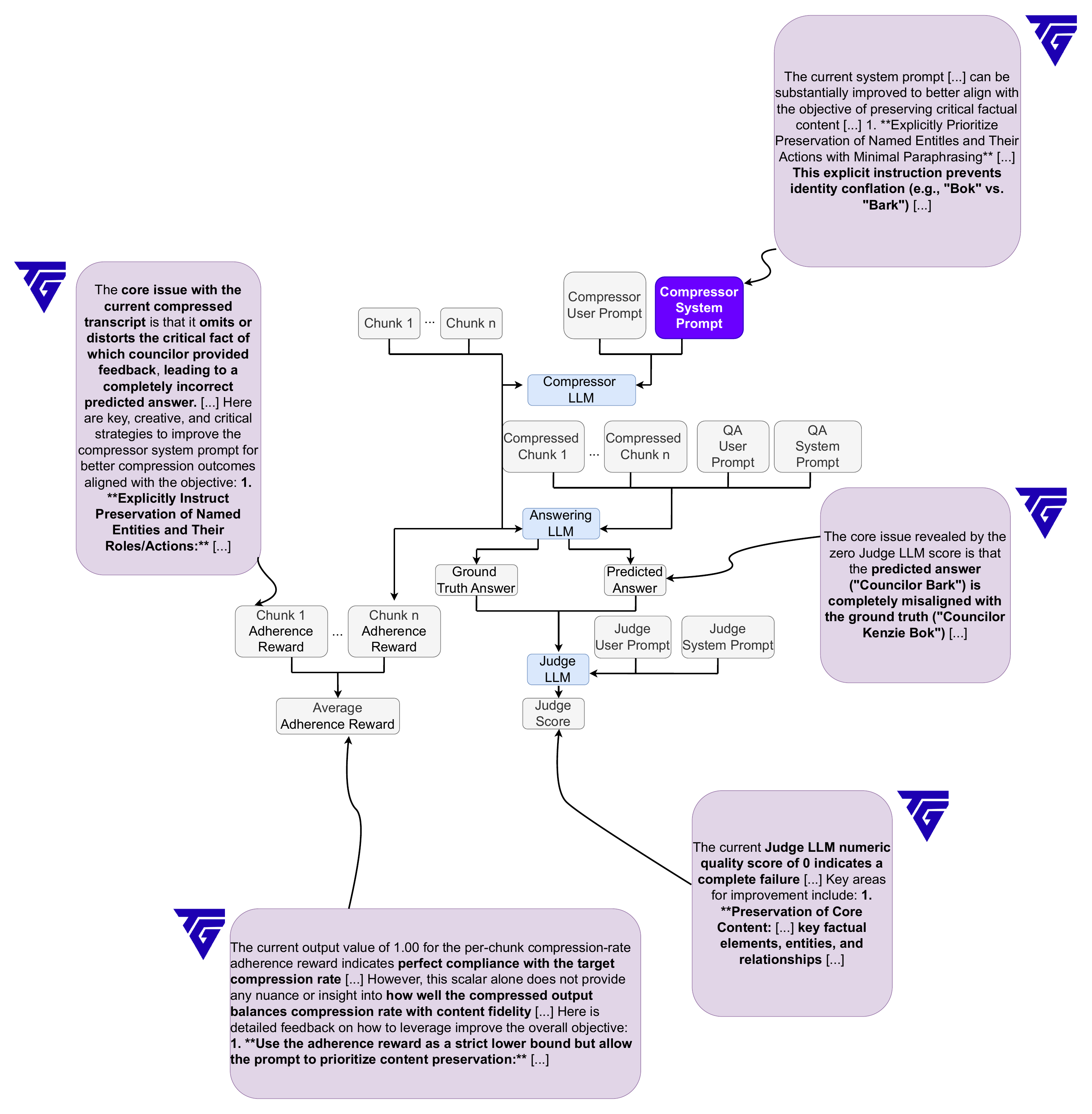}
  \caption{\textbf{\textsc{TextGrad} Computation Graph on the MeetingBank QA dataset}.}
  \label{fig:textgrad-graph}
\end{figure}

Figure \ref{fig:textgrad-graph} illustrates the computation graph involved in our prompt optimization procedure. The grey blocks represent text-based variables, the blue color denotes LLM nodes, and the optimized compressor system prompt variable is depicted in indigo. Meanwhile, the purple blocks show excerpts from \textsc{TextGrad} textual gradients. In this example, the natural language feedback generated by the optimizer LLM (i) identifies a question answering mistake (namely, the Answering LLM predicts ``Councilor Bark'' instead of the correct ``Councilor Bok'') and (ii) instructs the compressor to explicitly preserve named entities to prevent such mistakes in the future. 

The main goal of prompt optimization is to improve (i) downstream QA quality with compressed context, and (ii) adherence to a user-specified CR. 

\paragraph{QA quality scoring.} To assess information retention, we use synthetic QA pairs generated from the uncompressed MeetingBank transcript. Given a compressed transcript as context, the \emph{answering LLM} (gpt-4.1-mini) predicts an answer. Next, the \emph{judge LLM} (gpt-4.1-mini) scores this prediction against the ground truth answer on a discrete $0$ to $10$ scale.

\paragraph{CR Adherence scoring.} For a chunk with the target CR $r_{\text{tgt}}$ and the actual produced CR $r_{\text{act}} \!=\! \tfrac{N_{\text{comp}}}{N_{\text{orig}}}$, the per–chunk adherence reward $\mathrm{Adh}_{C}$ is given by:
\begin{equation}
\label{eq:adherence_reward}
\text{Adh}_{C} = 1 - \max(0, r_{\text{act}} - r_{\text{tgt}}), \quad \text{Adh}_{C} \in (-\infty, 1]
\end{equation}
Note that in the above formula $N_{\text{comp}}$ is the number of tokens in the produced compression and $N_{\text{orig}}$ is the number of tokens in the original input. 

For a transcript $T$ split into $K$ chunks $\{C_k\}_{k=1}^K$, we report the transcript–level CR adherence reward as the average of the rewards for each of its constituent chunks:
\begin{align}
\mathrm{Adh}_{T} \;=\; \frac{1}{K}\sum_{k=1}^{K} \mathrm{Adh}_{C_k}
\end{align}

\paragraph{Pseudocode overview.} We can now examine the \textsc{TextGrad} prompt optimization approach in greater detail. Overall, we can identify three main components of the algorithm.

\paragraph{1. Sampling and Compression}
\begin{enumerate}
  \item Sample a small batch of $TextGradBatchSize=2$ training transcripts.
  \item Split each transcript into chunks of at most $MaxChunkTokens=512$ tokens.
  \item For each chunk, draw a target CR $r_{\text{tgt}} \sim \mathcal{U}(0.1,\,0.7)$ and compute the token budget $N_{\text{tgt}}$ as follows:
  \[
  N_{\text{tgt}} \;=\; \bigl\lfloor r_{\text{tgt}} \cdot N_{\text{orig}} \bigr\rfloor 
  \]
  \item Prompt the LLM-based compressor to produce a compressed chunk under budget $N_{\text{tgt}}$.
  \item Concatenate the compressed chunks to obtain the compressed transcript.
\end{enumerate}

\paragraph{2. QA-based Evaluation}
\begin{enumerate}
  \item For each compressed transcript $t$, evaluate $Q=20$ synthetic questions and retain $k=2$ questions with the lowest judge scores. Let $H_t$ be the indices of these hard questions ($|H_t|=k$). The per-transcript judge average is:
  \begin{equation}
    J_t = \frac{1}{k} \sum_{i \in H_t} S_{t,i}
  \end{equation}
  where $S_{t,i}$ is the \textbf{judge score} for q $i$ in $t$, and $J_t$ is the \textbf{mean quality score} for that transcript.

  \item Aggregate metrics over batch size $B$. For a transcript $t$ split into $C_t$ chunks with per-chunk adherence $a_{t,j}$, the transcript adherence is:
  \begin{equation}
    A_t = \frac{1}{C_t} \sum_{j=1}^{C_t} a_{t,j}
  \end{equation}
  where $a_{t,j}$ is the \textbf{token-limit adherence} for chunk $j$, and $A_t$ is the \textbf{total adherence}.

  \item The final batch aggregates are:
  \begin{equation}
    \bar{J} = \frac{1}{B} \sum_{t=1}^{B} J_t, \qquad \bar{A} = \frac{1}{B} \sum_{t=1}^{B} A_t
  \end{equation}
\end{enumerate}

\paragraph{3. \textsc{TextGrad} Optimization Step}
\begin{enumerate}
  \item Use $\bar{J}$ and $\bar{A}$ as \textsc{TextGrad} optimization objectives to generate a new candidate prompt.
  \item Validate the candidate on $100$ hold-out transcripts.
  \item Accept the update if the new prompt increases either $\bar{J}$ or $\bar{A}$ when compared to the previous prompt.
\end{enumerate}

Combining the three key components of our prompt optimization approach, we present the complete algorithm in pseudocode section above.

\begin{algorithm}[ht]
\scriptsize
\caption{TextGrad Prompt Optimization via QA and Compression-Rate Adherence}
\label{alg:textgrad-qa}
\begin{algorithmic}[1]

\Require
  $InitialPrompt$ $\gets$ original system prompt for the compressor;\\
  $\mathcal{V}$: validation set of 100 MeetingBank transcripts;\\
  $\{\mathcal{B}_1,\ldots,\mathcal{B}_m\}$: training batches (size $=$ $TextgradBatchSize=2$);\\
  $SyntheticQAs$: mapping from transcript id $\to$ the list of question-answer tuples $(q,a)$;\\
  $AnswerLLM$: model used to answer questions (gpt-4.1-mini);\\
  $CompressorLLM$: abstractive compressor (its system prompt is optimized, Qwen3-4B);\\
  $JudgeLLM$: LLM judge scoring predicted vs.\ ground truth answer on the scale $[0,10]$ (gpt-4.1-mini);\\
  Hyperparameters: $MaxChunkTokens=512$, $NumHardQ=2$, target ratio range $[0.1,0.7]$.
\Ensure OptimizedPrompt
\Procedure{OptimizePrompt}{}
    \State $CurrentPrompt \gets InitialPrompt$
    \State $(PrevJudgeAvg, PrevAdherenceAvg) \gets \Call{MeasurePerformance}{\newline\mathcal{V}, CurrentPrompt}$ \Comment average judge score and average rate-adherence
    \For{each batch $\mathcal{B}$ in $\{\mathcal{B}_1,\ldots,\mathcal{B}_m\}$}
        \State $TrainingPoints \gets \emptyset$
        \For{each transcript $T$ in $\mathcal{B}$}
            \State $(CompressedT,~Adh_T) \gets \Call{CompressTranscript}{T, CompressorLLM, CurrentPrompt, \newline MaxChunkTokens}$
            \State $QAResults \gets \emptyset$
            \For{each $(q, a)$ in $SyntheticQAs[T.\text{id}]$}
                \State $pred \gets AnswerLLM(q,~\text{context}=CompressedT)$
                \State $score \gets JudgeLLM(pred,~q,~a)$ \Comment $0 \le score \le 10$
                \State $QAResults \gets QAResults \cup \{ \langle q, a, pred, score \rangle \}$
            \EndFor
            \State $Hardest \gets \Call{BottomK}{QAResults,~k=NumHardQ,~\text{by }score}$ \Comment lowest judge scores
            \For{each $r \in Hardest$}
                \State $TrainingPoints \gets TrainingPoints \cup \{ \langle r.q, r.a, r.pred, r.score, Adh_T, T.\text{id} \rangle \}$
            \EndFor
        \EndFor
        \State $CandidatePrompt \gets \Call{TextGradBackward}{TrainingPoints,~CurrentPrompt}$
        \State $(NewJudgeAvg, NewAdherenceAvg) \gets \Call{MeasurePerformance}{\mathcal{V},~CandidatePrompt}$
        \If{$(NewJudgeAvg > PrevJudgeAvg)~\lor~(NewAdherenceAvg > PrevAdherenceAvg)$}
            \State $CurrentPrompt \gets CandidatePrompt$
            \State $(PrevJudgeAvg, PrevAdherenceAvg) \gets (NewJudgeAvg, NewAdherenceAvg)$
        \EndIf
    \EndFor
    \State \Return $CurrentPrompt$ \Comment \textsc{OptimizedPrompt}
\EndProcedure
\end{algorithmic}
\end{algorithm}

\clearpage
% Requires: \usepackage{enumitem}
\begin{figure*}[t] % [t] places it at the top of the page
\begin{tcolorbox}[promptbox]
You are an agent tasked with compressing user prompts by preserving only the necessary information,
relationships, and required answer format to support accurate question answering. Remove all
unnecessary details and redundant information, and rephrase to shorten without losing \emph{any}
important content. Use aggressive shortening techniques—abbreviations, dropping
prepositions/articles, compact notation, and single-letter variables—only when clarity and
unambiguous understanding are maintained.

\textbf{Key instructions:}

\begin{itemize}[leftmargin=*, itemsep=0.5em]
  \item \textbf{Always preserve the full names
and official roles of all individuals mentioned, especially presenters, introducers, speakers,
decision-makers, and key actors linked to actions or sentiments.}
    \begin{itemize}[leftmargin=*, itemsep=0.3em]
      \item Never replace, omit, or
generalize person names with generic terms (e.g., “Council member,” “moderator”).
      \item This is
especially critical for any speaker who asks a question or makes a statement relevant to the
question(s).
      \item Example:
        \begin{itemize}[leftmargin=*, itemsep=0.2em]
          \item Good: “Jordan Win asked about next steps.”
          \item Bad: “A
council member asked about next steps.”
        \end{itemize}
    \end{itemize}

  \item \textbf{Always preserve explicit dates, times, and scheduling
information exactly as stated.}
    \begin{itemize}[leftmargin=*, itemsep=0.3em]
      \item If a date or time appears multiple times, it is acceptable to
keep or rephrase it concisely more than once to ensure clarity and completeness.
    \end{itemize}

  \item \textbf{Do not
paraphrase, abbreviate, or alter official titles, named individuals, or organizational names.}
    \begin{itemize}[leftmargin=*, itemsep=0.3em]
      \item Preserve all named entities and role titles verbatim, especially those linked to actions,
sentiments, or leadership roles.
      \item Maintain clear separation between titles and names (e.g.,
“Councilor Royal” not “Chair and Counsel Royal”).
    \end{itemize}

  \item \textbf{Given the known question(s) the compressed
text will support, prioritize retaining only those entities, facts, and relationships that directly
answer or support the question.}
    \begin{itemize}[leftmargin=*, itemsep=0.3em]
      \item Focus compression on preserving all entities, actions, and
relationships directly relevant to the question domain, especially procedural details and named-
entity attributions.
    \end{itemize}

  \item \textbf{Always preserve full names and official titles of individuals involved in
procedural actions such as motions, seconds, votes, and roll calls.}
    \begin{itemize}[leftmargin=*, itemsep=0.3em]
      \item Never omit, paraphrase,
or generalize these attributions.
      \item Maintain explicit speaker labels and preserve the sequence
of procedural actions to ensure clear attribution and temporal coherence.
      \item Example:
        \begin{itemize}[leftmargin=*, itemsep=0.2em]
          \item Good: “Motion to adopt by Councilman Smith, seconded by Councilman Ortega.”
          \item Bad: “Motion
adopted.”
        \end{itemize}
    \end{itemize}

  \item \textbf{Preserve explicit semantic relationships between entities and their attributes,
actions, votes, sentiments, or roles.}
    \begin{itemize}[leftmargin=*, itemsep=0.3em]
      \item Do not list entities without their associated relevant
facts.
      \item Replace pronouns or vague references with explicit names or roles (e.g., “Councilman
Brooks”) to maintain clarity and avoid ambiguity.
    \end{itemize}

  \item \textbf{Retain all expressions of gratitude,
mentoring acknowledgments, personal development, and related sentiments linked to named
individuals.}
    \begin{itemize}[leftmargin=*, itemsep=0.3em]
      \item These are critical for accurate question answering.
    \end{itemize}

  \item \textbf{Preserve causal and
definitional relationships linking entities, actions, and mandates, including temporal and
jurisdictional context relevant to the question.}

  \item \textbf{Maintain sufficient context around key
statements to preserve clear relationships between entities and their sentiments or roles, avoiding
isolated or ambiguous fragments.}
\end{itemize}
\end{tcolorbox}
\end{figure*}

\paragraph{Notable prompts.} We will now present selected prompts used in the \textsc{TextGrad} pipeline.

\paragraph{Compressor User Prompt.} Following \textsc{TextGrad}'s practice of reusing a shared system prompt across inputs, we specify the desired token budget in the user prompt. The user prompt is lean by design, as most of the compression instructions will be supplied in the system prompt.

\begin{tcolorbox}[promptbox]
  \scriptsize
  Please compress the text below. The length of the resulting compression \emph{must} be \{desired\_length\} tokens.\\
  Text to compress:
\end{tcolorbox}

\paragraph{Best Quality System Prompt.} The best-quality \textsc{TextGrad} prompt aggregates optimizer LLM strategies over multiple \textsc{TextGrad} iterations. It explicitly stresses the importance of named entities, numeric values, and other facts that are likely to appear in the downstream synthetic questions. The updated prompt also uses few-shot demonstrations. Through numerous examples and instructions, the updated prompt distills part of the optimizer LLM’s knowledge into the student Qwen3-4B compressor.

\begin{figure*}[ht] % [t] places it at the top of the page
\begin{tcolorbox}[promptbox]
\begin{itemize}[leftmargin=*, itemsep=0.5em]
  \item \textbf{Preserve all named individuals mentioned as representatives
or key actors, even if mentioned briefly or infrequently.}
    \begin{itemize}[leftmargin=*, itemsep=0.3em]
      \item Do not replace specific names with
generic entities unless explicitly stated.
    \end{itemize}

  \item \textbf{Preserve all numerical data exactly, including vote
counts, funding amounts, dates, and other quantitative details.}
    \begin{itemize}[leftmargin=*, itemsep=0.3em]
      \item Do not summarize or
approximate numbers.
    \end{itemize}

  \item \textbf{Avoid merging multiple speakers’ statements into generic summaries that
lose attribution or specificity.}
    \begin{itemize}[leftmargin=*, itemsep=0.3em]
      \item Maintain clear, unambiguous speaker labels for all
utterances, especially questions and key statements.
    \end{itemize}

  \item \textbf{Omit or condense procedural or
conversational details only if they do not contain critical facts or actors relevant to the
question.}

  \item \textbf{Avoid ellipses, ambiguous placeholders, or fragmented phrasing that obscure
meaning.}
    \begin{itemize}[leftmargin=*, itemsep=0.3em]
      \item Use concise, complete sentences with logical connectors to maintain coherence and
semantic clarity.
    \end{itemize}

  \item \textbf{Use abstractive summarization to rephrase content concisely but completely,
ensuring no key factual information, names, roles, sentiments, or dates are omitted or altered.}
    \begin{itemize}[leftmargin=*, itemsep=0.3em]
      \item Use compact notation and abbreviations only when clarity and unambiguous understanding are
maintained.
    \end{itemize}

  \item \textbf{Prioritize semantic completeness and clarity over strict length limits.}
    \begin{itemize}[leftmargin=*, itemsep=0.3em]
      \item It
is acceptable to exceed length constraints moderately to preserve critical facts, semantic roles,
sentiments, and temporal details.
    \end{itemize}

  \item \textbf{Stop compressing if clarity, semantic completeness, or
solvability is at risk.}

  \item \textbf{Before finalizing, perform a thorough self-check to verify that all
critical facts—dates, times, named entities, official titles, numerical data, actor-action pairs,
sentiments, and mandates—are present, unambiguous, and correctly represented.}
    \begin{itemize}[leftmargin=*, itemsep=0.3em]
      \item Reinsert or
rephrase any missing or altered critical information.
    \end{itemize}

  \item \textbf{Maintain consistent formatting and naming
conventions across all compressed chunks to ensure coherence and avoid contradictory or fragmented
information.}

  \item \textbf{Focus on preserving content related to council members’ expressions of
gratitude, mentoring, personal acknowledgments, historical figure mentions, and procedural
details.}
    \begin{itemize}[leftmargin=*, itemsep=0.3em]
      \item Use proxy keywords such as “gratitude,” “mentoring,” “thank,” “historical figure,”
“Councilmember [Name]” to guide focus without revealing answers.
    \end{itemize}
\end{itemize}

\textbf{Examples:}
\begin{itemize}[leftmargin=*, itemsep=0.3em]
  \item Good:
“Councilman Ortega seconded the motion.”
  \item Bad: “A council member seconded the motion.”
  \item Good:
“Councilmember Richardson thanked Councilwoman Gonzalez for mentoring.”
  \item Bad: “Councilwoman
Gonzalez thanked someone.”
  \item Good: “The working session was held on April 12th.”
  \item Bad: “The
working session date was not mentioned.”
  \item Good: “Motion withdrawn by Councilor Smith.”
  \item Bad:
“Motion withdrawn.”
  \item Good: “Funding increased by \$300,000 for FY 2020.”
  \item Bad: “Funding
increased by several thousand dollars.”
\end{itemize}

\textbf{Remember, the compressed transcript is the sole context for answering questions;
ensure it contains all information needed to answer temporal, factual, and named-entity questions
accurately.}
\end{tcolorbox}
\end{figure*}

\end{document}